\documentclass[english]{IEEEtran}
\usepackage[T1]{fontenc}
\usepackage[latin9]{inputenc}
\usepackage{array}
\usepackage{verbatim}
\usepackage{float}
\usepackage{wrapfig}
\usepackage{amsmath}
\usepackage{graphicx}
\usepackage{amssymb}

\makeatletter

\newcommand*\LyXbar{\rule[0.585ex]{1.2em}{0.25pt}}
\providecommand{\tabularnewline}{\\}
\floatstyle{ruled}
\newfloat{algorithm}{tbp}{loa}
\floatname{algorithm}{Algorithm}

\usepackage{url}

\makeatother

\usepackage{babel}

\begin{document}

\title{The path inference filter: model-based low-latency map matching of
probe vehicle data}

\author{Timothy Hunter, Pieter Abbeel, and Alexandre Bayen\\
University of California at Berkeley}
\maketitle
\begin{abstract}
We consider the problem of reconstructing vehicle trajectories from
sparse sequences of GPS points, for which the sampling interval is
between 10 seconds and 2 minutes. We introduce a new class of algorithms,
called altogether \emph{path inference filter} (PIF), that maps GPS
data in real time, for a variety of trade-offs and scenarios, and
with a high throughput. Numerous prior approaches in map-matching
can be shown to be special cases of the path inference filter presented
in this article. We present an efficient procedure for automatically
training the filter on new data, with or without ground truth observations.
The framework is evaluated on a large San Francisco taxi dataset and
is shown to improve upon the current state of the art. This filter
also provides insights about driving patterns of drivers. The path
inference filter has been deployed at an industrial scale inside the
\emph{Mobile Millennium }traffic information system, and is used to
map fleets of data in San Francisco, Sacramento, Stockholm and Porto.
\end{abstract}

\section{Introduction}

Amongst the modern man-made plagues, traffic congestion is a universally
recognized challenge \cite{downs2004still}. Building reliable and
cost-effective traffic monitoring systems is a prerequisite to addressing
this phenomenon. Historically, the estimation of traffic congestion
has been limited to highways, and has relied mostly on a static, dedicated
sensing infrastructure such as loop detectors or cameras \cite{Work08}.
The estimation problem is more challenging in the case of the secondary
road network, also called the \emph{arterial network}, due to the
cost of deploying a wide network of sensors in large metropolitan
areas. The most promising source of data is the GPS receiver in personal
smartphones and commercial fleet vehicles. According to some studies
\cite{schrank2009}, devices with a data connection and a GPS will
represent 80\% of the cellphone market by 2015. GPS observations in
cities are noisy \cite{cui2003autonomous}, and are usually provided
at low sampling rates (on the order of one minute) \cite{cabspotting}.
One of the common problems which occurs when dealing with GPS traces
is the correct mapping of these observations to the road network,
and the reconstruction of the trajectory of the vehicle. We present
a new class of algorithms, called the \emph{path inference filter,
}that solves this problem in a principled and efficient way. Specific
instantiations of this algorithm have been deployed as part of the
\emph{Mobile Millennium} system, which is a traffic estimation and
prediction system developed at the University of California \cite{mmfinalreport}.
\emph{Mobile Millennium} infers real-time traffic conditions using
GPS measurements from drivers running cell phone applications, taxicabs,
and other mobile and static data sources. This system was initially
deployed in the San Francisco Bay area and later expanded to other
locations such as Sacramento, Stockholm, and Porto.

\begin{figure}
\begin{centering}
\includegraphics[width=1\columnwidth]{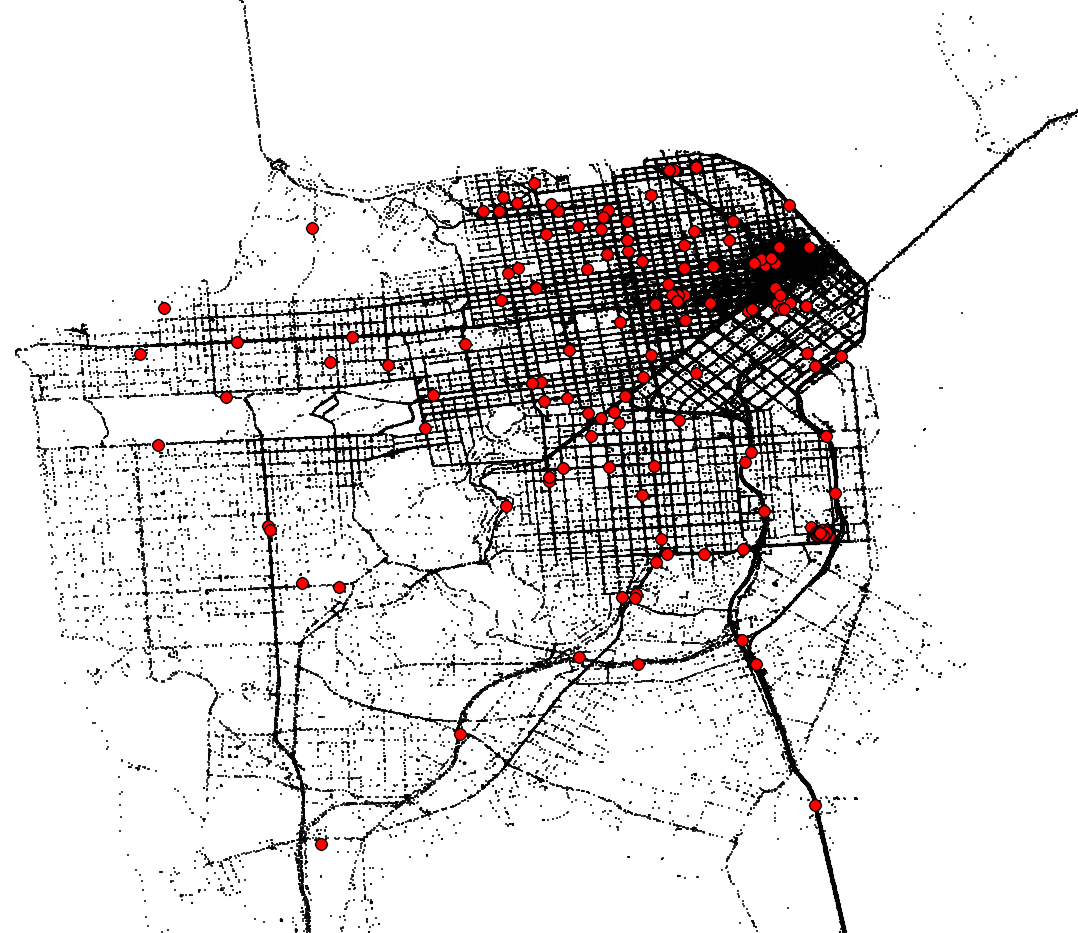}
\par\end{centering}

\caption{An example of dataset available to \emph{Mobile Millennium} and processed
by the path inference filter: taxicabs in San Francisco from the Cabspotting
program \cite{cabspotting}. Large circles in red show the position
of the taxis at a given time and small dots (in black) show past positions
(during the last five hours) of the fleet. The position of each vehicle
is observed every minute.\label{fig:mm_cloud_point}}
\end{figure}

GPS receivers have enjoyed a widespread use in transportation and
they are rapidly becoming a commodity. They offer unique capabilities
for tracking fleets of vehicles (for companies), and routing and navigation
(for individuals). These receivers are usually attached to a car or
a truck, also called a \emph{probe vehicle}, and they relay information
to a base station using the data channels of cellphone networks (3G,
4G). A typical datum provided by a probe vehicle includes an identifier
of the vehicle, a (noisy) position and a timestamp%
\footnote{The experiments in this article use GPS observations only. However,
nothing prevents the application of the algorithms presented in this
article to other types of localized data.%
}. Figure \ref{fig:mm_cloud_point} graphically presents a subset of
probe data collected by \emph{Mobile Millennium}. In addition to these
geolocalization attributes, data points contain other attributes such
as heading, speed, etc. We will show how this additional information
can be integrated in the rest of the framework presented in this article.

\begin{figure}
\begin{centering}
\includegraphics[width=5cm]{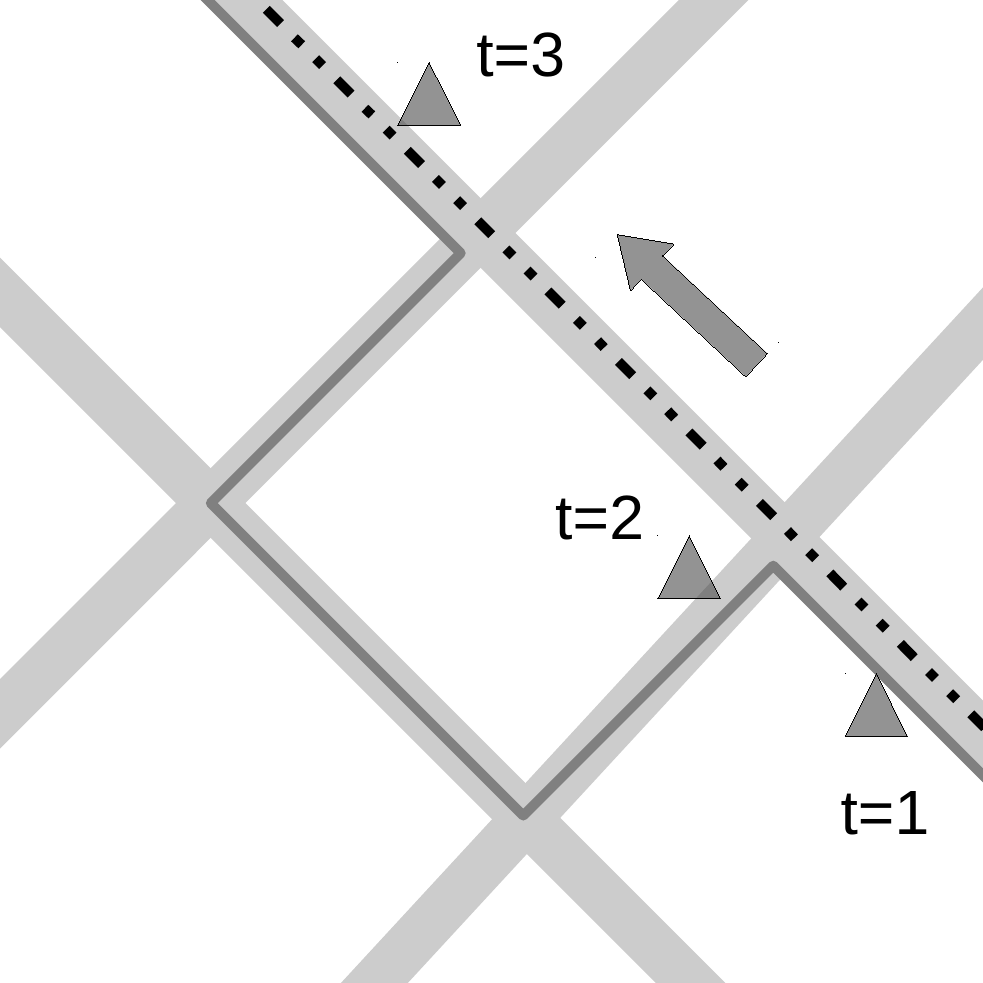}
\par\end{centering}

\caption{Example of failure when using an intuitive algorithm projects each
GPS measurement to the closest link. The raw GPS measurements are
the triangles, the actual true trajectory is the dashed line, and
the reconstructed trajectory is the continuous line. Due to noise
in the observation, the point at $t=2$ is closer to the orthogonal
road and forces the algorithm to add a left turn, while the vehicle
is actually going straight. This problem is frequently observed for
GPS data in cities. The \emph{path inference filter }provides one
solution to this problem.\label{fig:failure_shortest_proj}}
\end{figure}

\begin{figure}
\begin{centering}
\includegraphics[width=5cm]{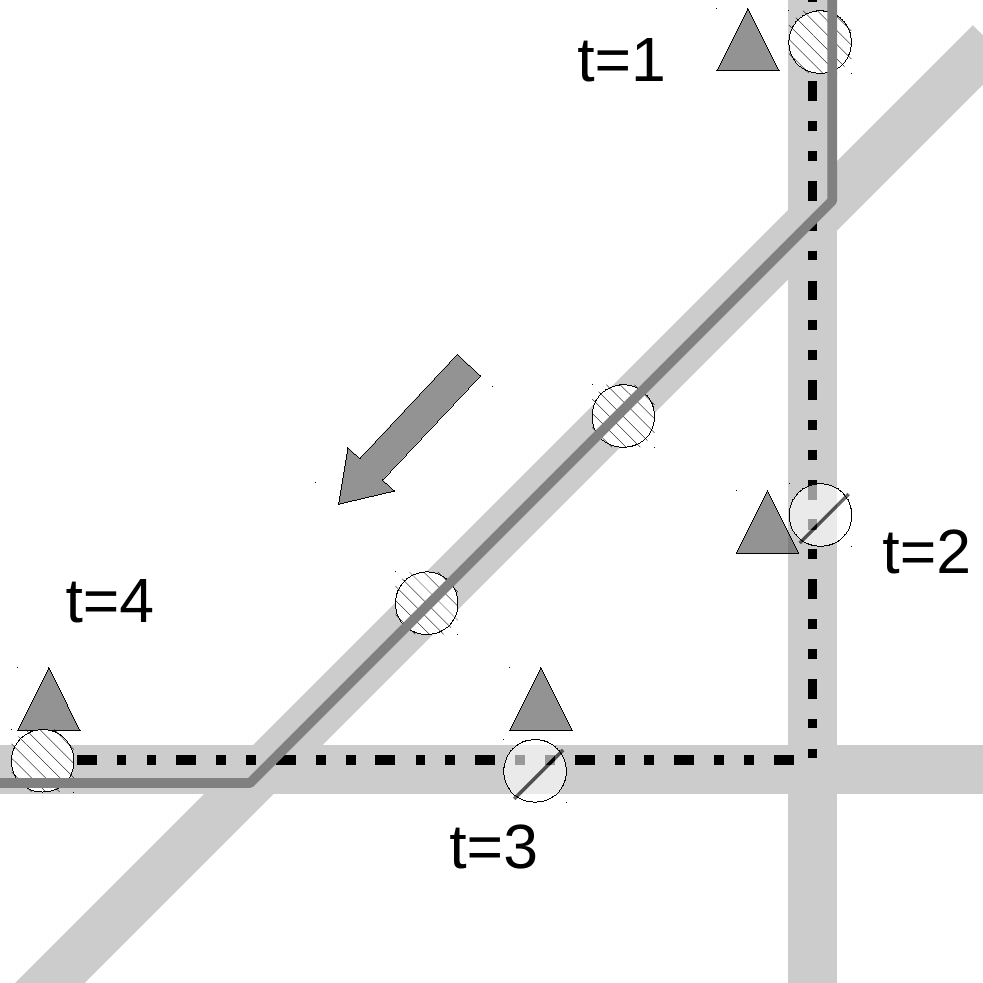}
\par\end{centering}

\caption{Example of failure when trying to minimize the path length between
a sequence of points. The raw observations are the triangles, the
actual true trajectory is the dashed line, and the reconstructed trajectory
is the continuous line. The circles are possible locations of the
vehicle corresponding to the observations. The hashed circles are
the states chosen by this reconstruction algorithm. Due to GPS errors
that induce problems explained in Figure \ref{fig:failure_shortest_proj},
we must consider point projections on all links within a certain distance
from the observed GPS points. However, the path computed by a shortest
path algorithm may not correspond to the true trajectory. Note how,
for $t=2$ and $t=3$, the wrong link and the wrong states are elected
to reconstruct the trajectory.\label{fig:failure_shortest_path}}
\end{figure}

The two most important characteristics of GPS data for traffic estimation
purposes are the GPS localization accuracy and the sampling strategy
followed by the probe vehicle. In order to reduce power consumption
or transmission costs, probe vehicles do not continuously report their
location to the base station. The probe data currently available are
generated using a combination of the two following strategies:
\begin{itemize}
\item \emph{Geographical sampling}: GPS probes are programmed to send information
in the vicinity of \emph{virtual landmarks} \cite{liu2008study}.
This concept was popularized by Nokia under the term \emph{Virtual
Trip Line} \cite{hoh_2008}. These landmarks are usually laid over
some predetermined route followed by drivers.
\item \emph{Temporal sampling}: GPS probes send their position at fixed
rate. The critical factor is then the \emph{temporal resolution }of
the probe data. A low temporal resolution carries some uncertainty
as to which trajectory was followed. A high temporal resolution gives
access to the complete and precise trajectory of the vehicle. However,
the device usually consumes more power and communication bandwidth.
\end{itemize}
In the case of a high temporal resolution (typically, a frequency
greater than an observation per second), some highly successful methods
have been developed for continuous estimation \cite{thrun2002probabilistic,miwa2004route,du2004lane}.
However, most data collected at large scale today is generated by
commercial fleet vehicles. It is primarily used for tracking the vehicles
and usually has a low temporal resolution (1 to 2 minutes) \cite{navteq,inrix,telenav,cabspotting}.
In the span of a minute, a vehicle in a city can cover several blocks.
Therefore, information on the precise path followed by the vehicle
is lost. Furthermore, due to GPS localization errors, recovering the
location of a vehicle that just sent an observation is a non trivial
task: there are usually several streets that could be compatible with
any given GPS observation. Simple deterministic algorithms to reconstruct
trajectories fail due to misprojection (Figure \ref{fig:failure_shortest_path})
or shortcuts (Figure \ref{fig:failure_shortest_proj}). Such shortcomings
have motivated our search for a principled approach that jointly considers
the mapping of observations to the network and the reconstruction
of the trajectory.

The problem of mapping data points onto a map can be traced back to
1980 \cite{bentley1980efficient}. Researchers started systematic
studies after the introduction of the GPS system to civilian applications
in the 1990s \cite{quddus2003general}. These early approaches followed
a \emph{geometric} perspective, associating each observation datum
to some point in the network \cite{white2000some}. Later, this projection
technique was refined to use more information such as heading and
road curvature. This greedy matching, however, leads to poor trajectory
reconstruction since it does not consider the path leading up to a
point \cite{yuan2010interactive}. New deterministic algorithms emerged
to directly match partial trajectories to the road by using the topology
of the network \cite{greenfeld2002matching} and topological metrics
based on the Fréchet distance \cite{brakatsoulas2005map,wenk2006addressing}.
These deterministic algorithms cannot readily cope with ambiguous
observations \cite{miwa2004route}, and were soon expanded into probabilistic
frameworks. A number of implementations were explored: particle filters
\cite{pyo2001development,gustafsson2002particle}, Kalman filters
\cite{ochieng2009map}, Hidden Markov Models \cite{bierlaire2011probabilistic},
and less mainstream approaches based on Fuzzy Logic and Belief Theory.

Two types of information are missing in a sequence of GPS readings:
the exact location of the vehicle on the road network when the observation
was emitted, and the path followed from the previous location to the
new location. These problems are correlated. The aforementioned approaches
focus on high-frequency sampling observations, for which the path
followed is extremely short (less than a few hundred meters, with
very few intersections). In this context, there is usually a dominant
path that starts from a well-defined point, and Bayesian filters accurately
reconstruct paths from observations \cite{ochieng2009map,thrun2002probabilistic,gustafsson2002particle}.
When sampling rates are lower and observed points are further apart,
however, a large number of paths are possible between two points.
Researchers have recently focused on efficiently identifying these
correct paths and have separated the joint problem of finding the
paths and finding the projections into two distinct problems. The
first problem is path identification and the second step is projection
matching \cite{zheng2011weight,bierlaire2011probabilistic,yuan2010interactive,giovannininovel,thiagarajan2009vtrack}.
Some interesting trajectories mixing points and paths that use a voting
scheme have also recently been proposed \cite{yuan2010interactive}.
Our filter aims at solving the two problems at the same time, by considering
a single unified notion of \emph{trajectory}. 

The \emph{path inference filter} is a probabilistic framework that
aims at recovering trajectories and road positions from low-frequency
probe data in real time, and in a computationally efficient manner.
As will be shown, the performance of the filter degrades gracefully%
{} as the sampling frequency decreases, and it can be tuned to different
scenarios (such as real time estimation with limited computing power
or offline, high accuracy estimation).

The filter is justified from the Bayesian perspective of the noisy
channel and falls into the general class of \emph{Conditional Random
Fields }\cite{lafferty2001conditional}. Our framework can be decomposed
into the following steps:
\begin{itemize}
\item \emph{Map matching}: each GPS measurement from the input is projected
onto a set of possible candidate states on the road network.
\item \emph{Path discovery:} admissible paths are computed between pairs
of candidate points on the road network.
\item \emph{Filtering}: probabilities are assigned to the paths and the
points using both a stochastic model for the vehicle dynamics and
probabilistic driver preferences learned from data.
\end{itemize}
According to the very exhaustive %
{} review by Quddus et al. \cite{quddus2007current}, most map-matching
approaches fall into one of the four categories:
\begin{enumerate}
\item {}``Geometric'' methods, which pick the closest matching point.
The distance metric itself is the subject of variations by different
authors.
\item {}``Weighted topological'' methods, which use connectivity information
between links and various ways to weight the different paths.
\item {}``Probabilistic'' methods, which combine variance information
about the points and topological information about the paths in a
simple way.
\item {}``Advanced'' methods, which encompass everything more complicated:
Kalman Filtering, Particle Filtering, Belief Theory \cite{el2005road}
and Fuzzy Logic \cite{syed2004fuzzy}.%
\footnote{Note that {}``probabilistic'' models, as well as most of the {}``advanced''
models (Kalman Filtering, Particle Filtering, Hidden Markov Models)
fall under the general umbrella of \emph{Dynamic Bayesian Filters},
presented in great detail in \cite{thrun2002probabilistic}. As such,
they deserve a common theoretical treatment, and in particular all
suffer from the same pitfalls detailed in Section \ref{sec:hmm}. %
}
\end{enumerate}
The path inference filter presents a number of compelling advantages
over the work found in the current literature:
\begin{enumerate}
\item The approach presents a general framework grounded in established
statistical theory that encompasses, as special cases, most techniques
presented as {}``geometric'', {}``topological'' or {}``probabilistic''.
In particular, it combines information about paths, points and network
topology in a single unified notion of \emph{trajectory.}
\item Nearly all work on Map Matching is segmented into (and presents results
for) either high-frequency or low-frequency sampling. The path inference
filter performs as well as the current state-of-the-art approaches
for sampling rates less than 30 seconds, and improves upon the state
of the art \cite{zheng2011weight,yuan2010interactive} by a factor
of more than 10\% for sampling intervals greater than 60 seconds%
\footnote{Performance comparisons are complicated by the lack of a single agreed-upon
benchmark dataset. Nevertheless, the city we study is complex enough
to compare favorably with cities studied with other works.%
}. We also analyze failure cases and we show that the output provided
by the path inference filter is always {}``close'' to the true output
for some metric.
\item As will be seen in Section \ref{sec:hmm}, most existing approaches
(which are based on Dynamic Bayesian Networks) do not work well at
lower frequencies due to the \emph{selection bias problem}. Our work
directly addresses this problem by performing inference on a Random
Field.
\item The path inference filter can be used with complex path models such
as those used in \cite{bierlaire2011probabilistic} and \cite{giovannininovel}.
In the present article, we restrict ourselves to a class of models
(the exponential family distributions) that is rich enough to provide
insight on the driving patterns of the vehicles. Furthermore, when
using this class of models, the learning of new parameters leads to
a convex problem formulation that is fast to solve. These parameters
can be learned using standard Machine Learning algorithms, even when
no ground truth is available.
\item With careful engineering, it is possible to achieve high throughput
on large-scale networks. Our reference implementation achieves an
average throughput of hundreds of GPS observations per second on a
single core in real time. Furthermore, the algorithm scales well on
multiple cores and has achieved average throughput of several thousands
of points per second on a multicore architecture.
\end{enumerate}
Algorithms often need to trade off accuracy for timeliness, and are
considered either {}``local'' (greedy) or {}``global'' (accumulating
some number of points before returning an answer) \cite{yuan2010interactive}.
The path inference filter is designed to work across the full spectrum
of accuracy versus latency. As we will show, we can still achieve
good accuracy by delaying computations by only one or two time steps.


\section{Path discovery}

The road network is described as a directed graph $\mathcal{N}=\left(\mathcal{V},\mathcal{E}\right)$
in which the nodes are the street intersections and the edges are
the streets, referred to in the text as the \emph{links }of the road
network. Each link is endowed with a number of physical attributes
(speed limit, number of lanes, type of road, etc.). Given a link of
the road network, the links into which a vehicle can travel will be
called \emph{outgoing links}, and the links from which it can come
will be called the \emph{incoming links.} Every location on the road
network is completely specified by a given link $l$ and offset $o$
on this link. The offset is a non-negative real number bounded by
the length of the corresponding link, and represents the position
on the link. At any time, the \emph{state }$x$ of a vehicle consists
of its location on the road network and some other optional information
such as speed, or heading. For our example we consider that the state
is simply the location on one of the road links (which are directed).
Additional information such as speed, heading, lane, etc. can easily
be incorporated into the state%
{}:\[
x=\left(l,o\right)\]

%
{}Furthermore, for the remainder of this article we consider trajectory
inference for a single probe vehicle.

\subsection{From GPS points to discrete vehicle states}

\label{sub:gps-to-states}
\begin{figure}
\begin{centering}
\includegraphics[width=1\columnwidth]{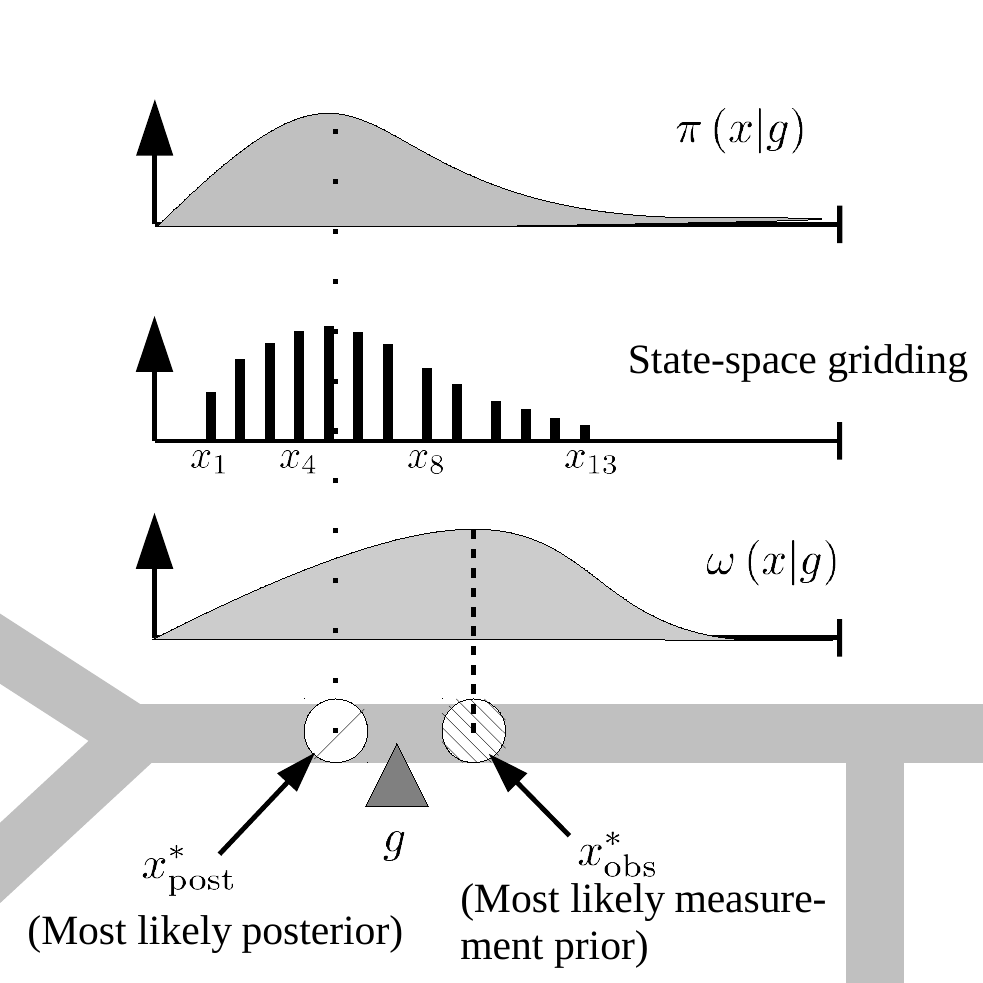}
\par\end{centering}

\caption{Example of a measurement $g$ on a link and two strategies to associate
state projections to that measurement on a particular link (gridding
and most likely location). The GPS measurement is the triangle denoted
$g$. For this particular measurement, the observation distribution
$\omega\left(x|g\right)$ and the posterior distribution $\pi\left(x|g\right)$
are also represented. When gridding, we select a number of states
$x_{1},\cdots x_{I}$ spanning each link at regular intervals. This
allows us to use the posterior distribution and have a more precise
distribution over the location of the vehicle. However, it is more
expensive to compute. Another strategy is to consider a single point
at the most probable offset $x_{\text{post}}^{*}$ according to the
posterior distribution $\pi\left(x|g\right)$. However, this location
depends on the prior, which is usually not available at this stage
(since the prior depends on the location of past and future points,
for which do not also know the location). A simple approximation is
to consider the most likely point $x_{\text{obs}}^{*}$according to
the observation distribution.\label{fig:Example-of-observation}}
\end{figure}

The points are mapped to the road following
a Bayesian formulation. Consider a GPS observation $g$. We study
the problem of mapping it to the road network according to our knowledge
of how this observation was generated. This generation process is
represented by a probability distribution $\omega\left(g|x\right)$
that, given a state $x$, returns a probability distribution over
all possible GPS observations $g$. Such distributions $\omega$ will
be described in Section \ref{sub:obs-mmodel}. Additionally, we may
have some \emph{prior knowledge} over the state of the vehicle. For
example, some links may be visited more often than others, and some
positions on links may be more frequent, such as when vehicles accumulate
at the intersections. This knowledge can be encoded in a \emph{prior
distribution }$\Omega\left(x\right)$. Under this general setting,
the state of a vehicle, given a GPS observation, can be computed using
Bayes' rule:\[
\pi\left(x|g\right)\propto\omega\left(g|x\right)\Omega\left(x\right)\]
The letter $\pi$ will define general probabilities, and their dependency
on variables will always be included. This probability distribution
is defined up to a scaling factor in order to integrate to~$1$.
This posterior distribution is usually complicated, owing to the mixed
nature of the state. The state space is the product of a discrete
space over the links and a continuous space over the link offsets.
Instead of representing it in closed form, some sampled values are
considered: for each link $l_{i}$, a finite number of states from
this link are elected to represent the posterior distribution of the
states on this link $\pi\left(o|g,l=l_{i}\right)$. A first way of
accomplishing this task is to grid the state space of each link, as
illustrated in Figure \ref{fig:Example-of-observation}. This strategy
is robust against the observation errors described in Section \ref{backward-paths},
but it introduces a large number of states to consider. Furthermore,
when new GPS values are observed every minute, the vehicle can move
quite extensively between updates. The grid step is usually small
compared to the distance traveled. Instead of defining a coarse grid
over each link, another approach is to use some \emph{most likely
state} on each link. Since our state is the pair of a link and an
offset on this link, this corresponds to selecting the most likely
offset on each state:\[
\forall l_{i},\;\; o_{i_{\text{posterior}}}^{*}=\underset{o}{\text{argmax}}\;\pi\left(o|g,l=l_{i}\right)\]

In practice, the probability distribution $\pi\left(x|g\right)$ decays
rapidly, and can be considered overwhelmingly small beyond a certain
distance from the observation \emph{$g$}. Links located beyond a
certain radius need not be considered valid projection links, and
may be discarded.

In the rest of this article, the boldface symbol $\mathbf{x}$ will
denote a (finite) collection of states associated with a GPS observation
$g$ that we will use to represent the posterior distribution $\pi\left(x|g\right)$,
and the integer $I$ will denote its cardinality: $\mathbf{x}=\left(x_{i}\right)_{1:I}$.
These points are called \emph{candidate state projections for the
GPS measurement $g$.} These discrete points will then be linked together
through trajectory information that takes into account the trajectory
and the dynamics of the vehicle. We now mention a few important points
for a practical implementation.

%
{}

\textbf{The prior distribution.} A Bayesian formulation requires that
we endow the state $x$ with a prior distribution $\Omega\left(x\right)$
that expresses our knowledge about the distribution of points on a
link. When no such information is available, since the offset is continuous
and bounded on a segment, a natural non-informative prior is the uniform
distribution over offsets: $\Omega\sim U\left(\left[0,\text{length}\left(l\right)\right]\right)$.
In this case, maximizing the posterior is equivalent to maximizing
the conditional distribution from the generative model: \[
\forall l_{i},o_{i_{\text{observation}}}^{*}=\text{argmax}_{o}\omega\left(g|x=\left(o,l_{i}\right)\right)\]

Having mapped GPS points into discrete points on the road network,
we now turn our attention to connecting these points by paths in order
to form trajectories.

\subsection{From discrete vehicle states to trajectories}

\begin{figure}
\begin{centering}
\includegraphics[width=0.9\columnwidth]{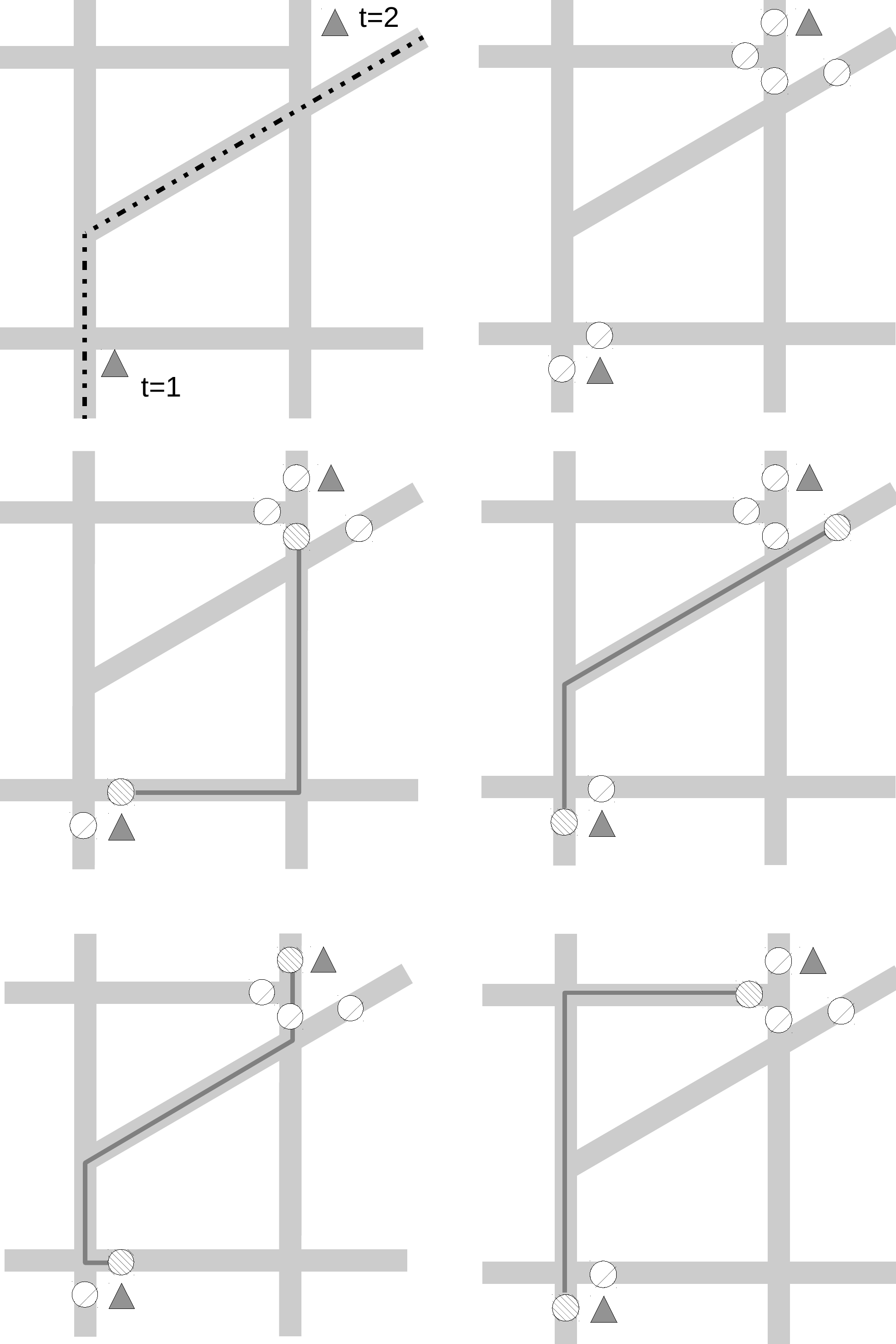}
\par\end{centering}

\caption{Example of path exploration between two observations. The true trajectory
and two associated GPS observations are shown on the upper left corner.
The upper right corner figure shows the set of candidate projections
associated with each observation. A path discovery algorithm computes
every acceptable path between between each pair of candidate projections.
The four figures at the bottom show a few examples of such computed
paths.\label{fig:path_exploration_example}}
\end{figure}

At each time step $t$, a GPS point $g^{t}$ (originating from a single
vehicle) is observed. This GPS point is then mapped onto $I^{t}$
different candidate states denoted $\mathbf{x}^{t}=x_{1}^{t}\cdots x_{I^{t}}^{t}$.
Because this set of projections is finite, there is only a (small)
finite number $J^{t}$ of paths that a vehicle can have taken while
moving from some state $x_{i}^{t}\in\mathbf{x}^{t}$ to some state
$x_{i'}^{t+1}\in\mathbf{x}^{t+1}$. We denote the set of \emph{candidate
paths }between the observation $g^{t}$ and the next observation $g^{t+1}$
by $\mathbf{p}^{t}$ :

\[
\mathbf{p}^{t}=\left(p_{j}^{t}\right)_{j=1:J^{t}}\]
Each path $p_{j}^{t}$ goes from one of the projection states $x_{i}^{t}$
of $g^{t}$ to a projection state $x_{i'}^{t+1}$ of $g^{t+1}$. There
may be multiple pairs of states to consider, and between each pair
of states, there are typically several paths available (see Figure
\ref{fig:path_exploration_example}). Lastly, a \emph{trajectory}
is defined by the succession of states and paths, starting and ending
with a state:\[
\tau=x_{1}p_{1}x_{2}\cdots p_{t-1}x_{t}\]

where $x_{1}$ is one element of $\mathbf{x}^{1}$, $p_{1}$ of $\mathbf{p}^{1}$,
and so on.

Due to speed limits leading to lower bounds on achievable travel times
on the network, there is only a finite number of paths a vehicle can
take during a time interval $\Delta t$. Such paths can be computed
using standard graph search algorithms. The depth of the search is
bounded by the maximum distance a vehicle can travel on the network
at a speed $v_{\text{max}}$ within the time interval between each
observation. An algorithm that performs well in practice is the A{*}
algorithm \cite{hart1968formal}, a common graph search algorithm
that makes use of a heuristic to guide its search. The cost metric
we use here is the expected travel time on each link, and the heuristic
is the shortest geographical distance, properly scaled so that it
is an admissible heuristic.

\begin{figure}
\begin{centering}
\includegraphics[width=6cm]{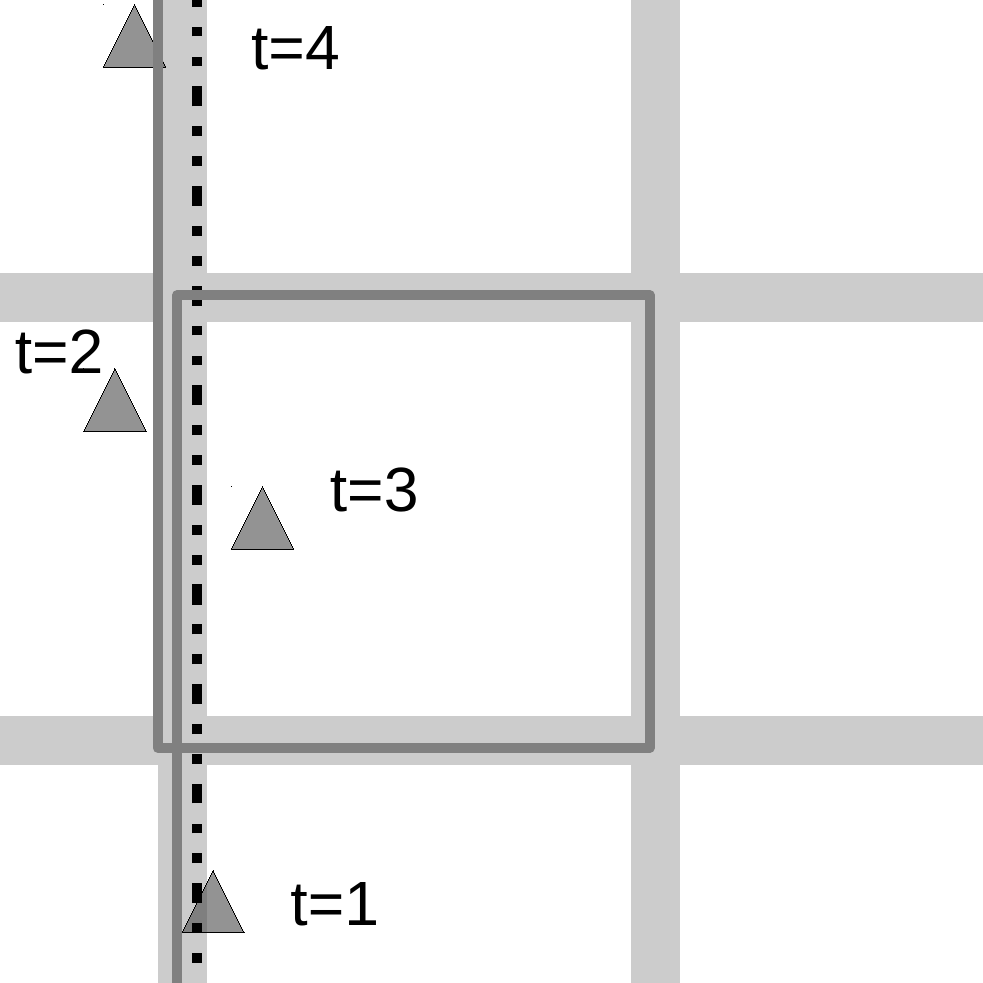}
\par\end{centering}

\caption{Example of failure when observing strict physical consistency: due
to the observation noise, the observation (3) appears physically behind
(2) on the same link. Without considering backward paths, the most
plausible explanation is that the vehicle performed a complete loop
around the neighboring block.\label{fig:failure_loop}}
\end{figure}

\textbf{\label{backward-paths}The case of backward paths.} It is
convenient and realistic to assume that a vehicle always drives \emph{forward},
i.e. in the same direction of a link%
\footnote{Reverse driving is in some cases even illegal. For example, the laws
of Glendale, Arizona, prohibit reverse driving.%
}. In our notation, a vehicle enters a link at offset $0$, drives
along the link following a non-decreasing offset, and exits the link
when the offset value reaches the total length of the link. However,
due to GPS noise, the most likely state projection of a vehicle waiting
at a red light may appear to go backward, as shown in Figure \ref{fig:failure_loop}.
This leads to incorrect transitions if we assume that paths only go
forward on a link. Three approaches to solve this issue are discussed,
depending on the application:

%
{}

\begin{enumerate}
\item It is possible to keep a single state for each link (the most likely)
and explore some backward paths. These paths are assumed to go backward
because of observation noise. This solution provides \emph{connected
states at the expense of physical consistency}: all the measurements
are correctly mapped to their most likely location, but the trajectories
themselves are not physically acceptable. This is useful for applications
that do not require connectedness between pairs of states, for example
when computing a distribution of the density of probe data per link.
\item It is also possible to disallow backward paths and consider multiple
states per link, such as a grid over the state space. A vehicle never
goes backward, and in this case the filter can generally account for
the vehicle not moving by associating the same state to successive
observations. All the trajectories are physically consistent and the
posterior state density is the same as the probability density of
the most likely states, but is more burdensome from a computational
perspective (the number of paths to consider grows quadratically with
the number of states).
\item Finally it is possible to disallow backward paths and use a sparse
number of states. The path connectivity issue is solved using some
heuristics. Our implementation creates a new state projection on a
link $l$ using the following approach:\\
Given a new observation $g$, and its most likely state projection
$x^{*}=\left(l,o^{*}\right)$:

\begin{enumerate}
\item If no projection for the link $l$ was found at the previous time
step, return $x^{*}$
\item If such a projection $x_{\text{before}}=\left(l,o_{\text{before}}\right)$
existed, return $x=\left(l,\max\left(o_{\text{before}},o^{*}\right)\right)$
\end{enumerate}
With this heuristic, all the points will be well connected, but the
density of the states will not be the same as the density of the most
likely reconstructed states.

\end{enumerate}
In summary, the first solution is better for density estimations
and the third approach works better for travel time estimations. The
second option is currently only used for high-frequency offline filtering,
for which paths are short, and for which more expensive computations
is an acceptable cost.

\textbf{Handling errors} Maps may contains some inaccuracies, and
may not cover all the possible driving patterns. Two errors were found
to have a serious effect on the performance of the filter:
\begin{itemize}
\item Out of network driving: This usually occurs in parking lots or commercials
driveways.
\item Topological errors: Some links may be missing on the base map, or
one-way streets may have changed to two-way streets. These situations
are handled by running \emph{flow analysis }on the trajectory graph\emph{.}
For every new incoming GPS point, after computing the paths and states,
it is checked if any candidate position of the first point of the
trajectory is reachable from any reachable candidate position on the
latest incoming point, or equivalently if the trajectory graph has
a positive flow. The set of state projections of an observation may
end up being disconnected from the start point even if at every step,
there exists a set of paths between each points. In this situation,
the probability model will return a probability of 0 (non-physical
trajectories) for any trajectory. If a point becomes unreachable from
the start point, the trajectory is broken, and restarted again from
this point. Trajectory breaks were few (less than a dozen for our
dataset), and a visual inspection showed that the vehicle was not
following the topology of the network and instead made U-turns or
breached through one-way streets.
\end{itemize}

\section{Discrete filtering using a Conditional Random Field}

\label{sec:hmm}

In the previous section, we reduced the trajectory reconstruction
problem to a discrete selection problem between sets of candidate
projection points, interleaved with sets of candidate paths. A probabilistic
framework can now be applied to infer a reconstructed trajectory $\tau$
or probability distributions over candidate candidate states and candidate
paths. Without further assumptions, one would have to enumerate and
compute probabilities for every possible trajectory. This is not possible
for long sequences of observations, as the number of possible trajectories
grows exponentially with the number of observations chained together.
By assuming additional independence relations, we turn this intractable
inference problem into a tractable one.

\subsection{Conditional Random Fields to weight trajectories}

The observation model provides the joint distribution of a state on
the road network given an observation. We have described the \emph{noisy
generative model }for the observations in Section \ref{sub:gps-to-states}.
Assuming that the vehicle is at a point $x$, a GPS observation $g$
will be observed according to a model $\omega$ that describes a noisy
observation channel. The value of $g$ only depends on the state of
the vehicle, i.e. the model reads $\omega\left(g|x\right)$. For every
time step $t$, assuming that the vehicle is at the location $x^{t}$,
a GPS observation $g^{t}$ is created according to the distribution
$\omega\left(g^{t}|x^{t}\right)$. 

Additionally, we endow the set of all possible paths on the road network
with a probability distribution. The \emph{transition model} $\eta$
describes the preference of a driver for a particular path. In probabilistic
terms, it provides a distribution $\eta\left(p\right)$ defined over
all possible paths $p$ across the road network. This distribution
is not a distribution over actually observed paths as much as a model
of the \emph{preferences }of the driver when given the choice between
several options.

We introduce the following \emph{Markov assumptions}.
\begin{itemize}
\item Given a start state $x_{\text{start}}$ and an end state $x_{\text{end}}$,
the path $p$ followed by the vehicle between these two points will
only depend on the start state, the end state and the path itself.
In particular, it will not depend on previous paths or future paths.
\item Consider a state $x$ followed by a path $p_{\text{next}}$ and preceded
by a path $p_{\text{previous}}$, and associated to a GPS measurement
$g$. Then the paths taken by the vehicle are independent from the
GPS measurement $g$ if the state $x$ is known. In other words, the
GPS measurement does not add subsequent information given the knowledge
of the state of the vehicle.
\end{itemize}
Since a state is composed of an offset and a link, a path is completely
determined by a start state, an end state and a list of links in between.
Conditional on the start state and end state, the number of paths
between these points is finite (it is the number of link paths that
join the start link and the end link).

Not every path is compatible with given start point and end point:
the path must start at the start state and must end at the end state.
We formally express the compatibility between a state $x$ and the
start state of a path $p$ with the compatibility function $\underline{\delta}$:

\[
\underline{\delta}\left(x,p\right)=\begin{cases}
1 & \text{if the path }p\text{ starts at point }x\\
0 & \text{otherwise}\end{cases}\]

Similarly, we introduce the compatibility function $\bar{\delta}$
to express the agreement between a state and the end state of a path:\[
\bar{\delta}\left(p,x\right)=\begin{cases}
1 & \text{if the path }p\text{ ends at point }x\\
0 & \text{otherwise}\end{cases}\]

Given a sequence of observations $g^{1:T}=g^{1}\cdots g^{T}$ and
an associated trajectory $\tau=x^{1}p^{1}\cdots x^{T}$, we define
the \emph{unnormalized score, }or \emph{potential},\emph{ }of the
trajectory as:\begin{eqnarray*}
\phi\left(\tau|g^{1:T}\right) & = & \left[\prod_{t=1}^{T-1}\omega\left(g^{t}|x^{t}\right)\underline{\delta}\left(x^{t},p^{t}\right)\eta\left(p^{t}\right)\bar{\delta}\left(p^{t},x^{t+1}\right)\right]\\
 &  & \cdot\omega\left(g^{T}|x^{T}\right)\end{eqnarray*}

The non-negative function $\phi$ is called the \emph{potential function.}
A trajectory $\tau$ is said to be a \emph{compatible trajectory with
the observation sequence }$g^{1:T}$ if $\phi\left(\tau|g^{1:T}\right)>0$.
When properly scaled, the potential $\phi$ defines a probability
distribution over all possible trajectories, given a sequence of observations:\[
\pi\left(\tau|g^{1:T}\right)=\frac{\phi\left(\tau|g^{1:T}\right)}{Z}\]
The variable $Z$, called the \emph{partition function}, is the sum
of the potentials over all the compatible trajectories:\[
Z=\sum_{\tau}\phi\left(\tau|g^{1:T}\right)\]

\begin{figure}
\begin{centering}
\includegraphics[width=1\columnwidth]{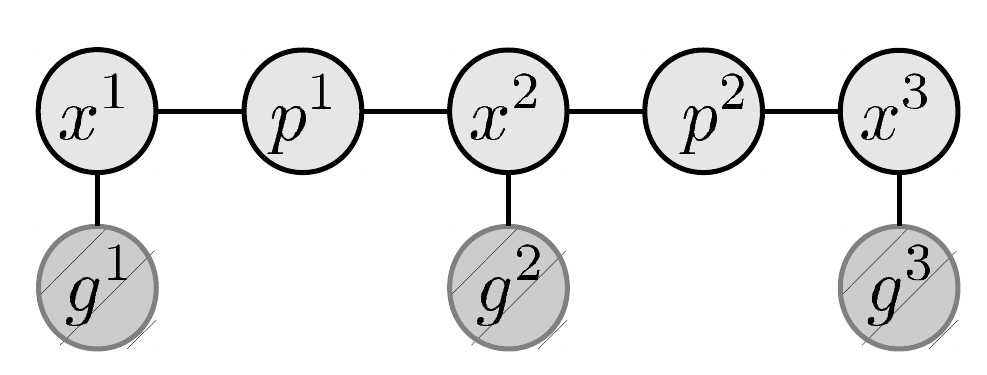}
\par\end{centering}

\caption{Illustration of the Conditional Random Field defined over a trajectory
$\tau=x^{1}p^{1}x^{2}p^{2}x^{3}$ and a sequence of observations $g^{1:3}$.
The gray nodes indicate the observed values. The solid lines indicate
the factors between the variables: $\omega\left(g^{t}|x^{t}\right)$
between a state $x^{t}$ and an observation $g^{t}$, $\underline{\delta}\left(x^{t},p^{t}\right)\eta\left(p^{t}\right)$
between a state $x^{t}$ and a path $p^{t}$ and $\bar{\delta}\left(p^{t},x^{t+1}\right)$
between a path $p^{t}$ and a subsequent state $x^{t+1}$.\label{fig:crf_model}}
\end{figure}

We have combined the observation model $\omega$ and the transition
model $\eta$ into a single potential function $\phi$, which defines
an unnormalized distribution over all trajectories. Such a probabilistic
framework is called a \emph{Conditional Random Field }(CRF) \cite{lafferty2001conditional}.
A CRF is an undirected graphical model which is defined as the unnormalized
product of factors over cliques of factors (see Figure \ref{fig:crf_model}).
There can be an exponentially large number of paths, so the partition
function cannot be computed by simply summing the value of $\phi$
over every possible trajectory. As will be seen in Section \ref{sec:training},
the value of $Z$ needs to be computed only during the training phase.
Furthermore it can be computed efficiently using dynamic programming.

\begin{figure}
\begin{centering}
\includegraphics[width=1\columnwidth]{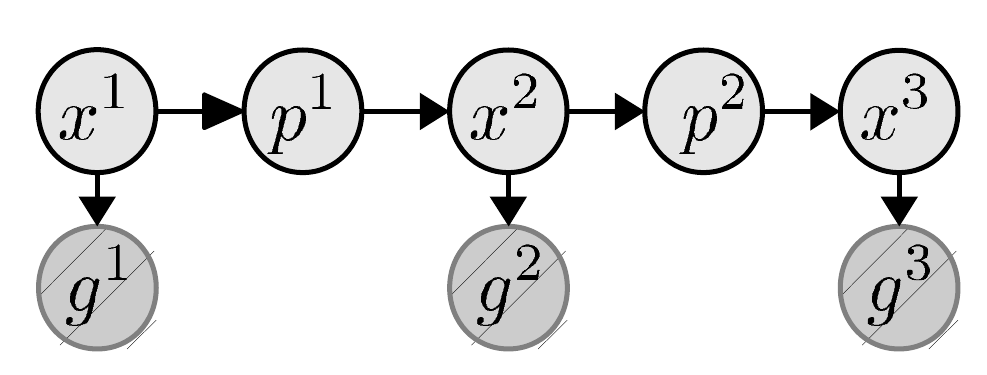}
\par\end{centering}

\caption{A Dynamic Bayesian Network (DBN) commonly used to model the trajectory
reconstruction problem. The arrows indicated the directed dependencies
of the variables. The GPS observations $g^{t}$ are generated from
states $x^{t}$. The unobserved paths $p^{t}$ are generated from
a state $x^{t}$, following a transition probability distribution
$\hat{\pi}\left(p|x\right)$. The transition from a path $p^{t}$
to a state $x^{t}$ follows the transition model $\hat{\pi}\left(x|p\right)$.
\label{fig:dbn_model}}
\end{figure}

\textbf{The case against the Hidden Markov Model approach}. The classical
approach to filtering in the context of trajectories is based on Hidden
Markov Models (HMMs), or their generalization, Dynamic Bayesian Networks
(DBNs)~\cite{murphy2002dynamic}: a sequence of states and trajectories
form a trajectory, and the coupling of trajectories and states is
done using transition models $\hat{\pi}\left(x|p\right)$ and $\hat{\pi}\left(p|x\right)$.
See Figure \ref{fig:dbn_model} for a representation.

This results in a chain-structured directed probabilistic graphical
model in which the path variables $p^{t}$ are unobserved. Depending
on the specifics of the transition models, $\hat{\pi}\left(x|p\right)$
and $\hat{\pi}\left(p|x\right)$, probabilistic inference has been
done with Kalman filters \cite{ochieng2009map,pyo2001development},
the forward algorithm or the Viterbi algorithm \cite{bierlaire2011probabilistic,bierlaire2008route},
or particle filters \cite{gustafsson2002particle}.

\begin{figure}
\begin{centering}
\includegraphics[width=6cm]{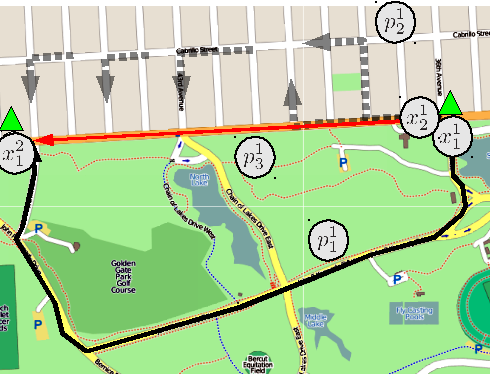}
\par\end{centering}

\caption{Example of a failure case when using a Hidden Markov Model: the solid
black path will be favored over all the other paths.\label{fig:hmm_failure}}
\end{figure}

Hidden Markov Model representations, however, suffer from the \emph{selection
bias problem}, first noted in the labeling of words sequences \cite{lafferty2001conditional},
which makes them not the best fit for solving path inference problems.
Consider the example trajectory $\tau=x^{1}p^{1}x^{2}$ observed in
our data, represented in Figure \ref{fig:hmm_failure}. For clarity,
we consider only two states $x_{1}^{1}$ and $x_{2}^{1}$ associated
with the GPS reading $g^{1}$ and a single state $x_{1}^{2}$ associated
with $g^{2}$. The paths $\left(p_{j}^{1}\right)_{j}$ between $x^{1}$
and $x^{2}$ may either be the lone path $p_{1}^{1}$ from $x_{1}^{1}$
to $x_{1}^{2}$ that allows a vehicle to cross the Golden Gate Park,
or one of the many paths between Cabrillo Street and Fulton Street
that go from $x_{2}^{1}$ to $x^{1}$, including $p_{3}^{1}$ and
$p_{2}^{1}$. In the HMM representation, the transition probabilities
must sum to 1 when conditioned on a starting point. Since there is
a single path from $x_{2}^{1}$ to $x^{2}$, the probability of taking
this path from the state $x_{1}^{1}$ will be $\hat{\pi}\left(p_{1}^{1}|x_{1}^{1}\right)=1$
so the overall probability of this path is $\hat{\pi}\left(p_{1}^{1}|g^{1}\right)=\hat{\pi}\left(x_{1}^{1}|g^{1}\right)$.
Consider now the paths from $x_{2}^{1}$ to $x_{1}^{2}$: a lot of
these paths will have a similar weight, since they correspond to different
turns and across the lattice of streets. For each path $p$ amongst
these $N$ paths of similar weight, Bayes' assumption implies $\hat{\pi}\left(p|x_{2}^{1}\right)\approx\frac{1}{N}$
so the overall probability of this path is $\hat{\pi}\left(p|g^{1}\right)\approx\frac{1}{N}\hat{\pi}\left(x_{2}^{1}|g^{1}\right)$.
In this case, $N$ can be large enough that $\hat{\pi}\left(p_{1}^{1}|g^{1}\right)\geq\hat{\pi}\left(p|g^{1}\right)$,
and the remote path will be selected as the most likely path.

Due to their structures, all HMM models will be biased towards states
that have the least expansions. In the case of a road network, this
can be pathological. In particular, the HMM assumption will carry
the effect of the selection bias as long as there are long disconnected
segments of road. This can be particularly troublesome in the case
of road networks since HMM models will end up being forced to assign
too much weight to a highway (which is highly disconnected) and not
enough to the road network alongside the highway. Our model, which
is based on Conditional Random Fields, does not have this problem
since the renormalization happens just once and is over all paths
from start to end, rather than renormalizing for every single state
transition independently.

\textbf{Efficient filtering algorithms.} Using the probabilistic framework
of the CRF, we wish to infer:
\begin{itemize}
\item the most likely trajectory $\tau$:\begin{equation}
\tau^{*}=\underset{\tau}{\text{argmax}}\;\;\pi\left(\tau|g^{1:T}\right)\label{eq:most-likely-traj}\end{equation}

\item the posterior distributions over the elements of the trajectory, i.e.
the conditional marginals $\pi\left(x^{t}|g^{1:T}\right)$ and $\pi\left(p^{t}|g^{1:T}\right)$
\end{itemize}
As will be seen, both elements can be computed without having to obtain
the partition function $Z$. The solution to both problems is a particular
case of the \emph{Junction Tree algorithm} \cite{murphy2002dynamic}
and can be computed in time complexity linear in the time horizon
by using a dynamic programing formulation. Computing the most likely
trajectory is a particular instantiation of a standard dynamic programing
algorithm called the \emph{Viterbi algorithm} \cite{forney1973viterbi}.
Using a classic Machine Learning algorithm for chain-structured junction
trees (the \emph{forward-backward algorithm} \cite{seymore1999learning,bilmes1998gentle}),
all the conditional marginals can be computed in two passes over the
variables. In the next section, we detail the justification for the
Viterbi algorithm and in Section \ref{sub:filtering} we describe
an efficient implementation of the forward-backward algorithm in the
context of this application.

\subsection{Finding the most likely path}

For the rest of this section, we fix a sequence of observations $g^{1:T}$.
For each observation $g^{t}$, we consider a set of candidate state
projections $\mathbf{x}^{t}$. At each time step $t\in\left[1\cdots T-1\right]$,
we consider a set of paths $\mathbf{p}^{t}$, so that each path $p^{t}$
from $\mathbf{p}^{t}$ starts from some state $x^{t}\in\mathbf{x}^{t}$
and ends at some state $x^{t+1}\in\mathbf{x}^{t+1}$. We will consider
the set $\varsigma$ of valid trajectories in the Cartesian space
defined by these projections and these paths:\[
\varsigma=\left\{ \tau=x^{1}p^{1}\cdots p^{T-1}x^{T}|\begin{array}{c}
x^{t}\in\mathbf{x}^{t}\\
p^{t}\in\mathbf{p}^{t}\\
\bar{\delta}\left(x^{t},p^{t}\right)=1\\
\underline{\delta}\left(p^{t},x^{t+1}\right)=1\end{array}\right\} \]
 In particular, if $I^{t}$ is the number of candidate states associated
with $g^{t}$ (i.e. the cardinal of $\mathbf{x}^{t}$) and $J^{t}$
is the number of candidate paths in $\mathbf{p}^{t}$, then there
are at most $\prod_{1}^{T}I^{t}\prod_{1}^{T-1}J^{t}$ possible trajectories
to consider. We will see, however, that most likely trajectory $\tau^{*}$
can be computed in $O\left(TI^{*}J^{*}\right)$ computations, with
$I^{*}=\max_{t}I^{t}$ and $J^{*}=\max_{t}J^{t}$.

The partition function $Z$ does not depend on the current trajectory
$\tau$ and need not be computed when solving Equation~\ref{eq:most-likely-traj}:\[
\begin{array}{cccc}
\tau^{*} & = & \underset{\tau\in\varsigma}{\text{argmax}} & \pi\left(\tau|g^{1:T}\right)\\
 & = & \underset{\tau\in\varsigma}{\text{argmax}} & \phi\left(\tau|g^{1:T}\right)\end{array}\]

Call $\phi^{*}\left(g^{1:T}\right)$ the maximum value over all the
potentials of the trajectories compatible with the observations $g^{1:T}$:\[
\phi^{*}\left(g^{1:T}\right)=\max_{\tau\in\varsigma}\phi\left(\tau|g^{1:T}\right)\]

The trajectory $\tau$ that realizes this maximum value is found by
tracing back the computations. For example, some pointers to the intermediate
partial trajectories can be stored to trace back the complete trajectory,
as done in the referring implementation \cite{pythonimpl}. This is
why we will only consider the computation of this maximum. The function
$\phi$ depends on the probability distributions $\omega$ and $\eta$,
left undefined so far. These distributions will be presented in depth
in Sections \ref{sub:obs-mmodel} and \ref{sub:transition}. 

It is useful to introduce notation related to a \emph{partial trajectory}.
Call $\tau^{1:t}$ the \emph{partial trajectory} until time step $t$:
\[
\tau^{1:t}=x^{1}p^{1}\cdots x^{t}\]
For a partial trajectory, we define some partial potentials $\phi\left(\tau^{1:t}|g^{1:t}\right)$
that depend only on the observations seen so far:\begin{eqnarray}
\phi\left(\tau^{1:t}|g^{1:t}\right) & = & \omega\left(g^{1}|x^{1}\right)\prod_{t'=1}^{t-1}\underline{\delta}\left(x^{t'},p^{t'}\right)\eta\left(p^{t'}\right)\nonumber \\
 &  & \cdot\bar{\delta}\left(p^{t'},x^{t'+1}\right)\omega\left(g^{t'+1}|x^{t'+1}\right)\label{eq:def-partial-phi}\end{eqnarray}

For each time step $t$, given a state index $i\in\left[1,I^{t}\right]$
we introduce the potential function for trajectories that end at the
state $x_{i}^{t}$:\[
\phi_{i}^{t}=\max_{\tau^{1:t}=x^{1}p^{1}\cdots.x^{t-1}p^{t-1}x_{i}^{t}}\phi\left(\tau^{1:t}|g^{1:t}\right)\]
One sees:\[
\phi^{*}=\max_{i\in\left[1,I^{T}\right]}\phi_{i}^{T}\]
The partial potentials defined in Equation \eqref{eq:def-partial-phi}
follow an inductive identity:

\[
\phi_{i}^{1}=\omega\left(g^{1}|x_{i}^{t}\right)\]
\begin{equation}
\forall t,\phi_{i}^{t+1}=\max_{\begin{array}{c}
i'\in\left[1,I^{t}\right]\\
j\in\left[1,J^{t}\right]\end{array}}\begin{array}{l}
\left[\phi_{i'}^{t}\underline{\delta}\left(x_{i'}^{t},p_{j}^{t}\right)\eta\left(p_{j}^{t}\right)\right.\\
\left.\cdot\bar{\delta}\left(p_{j}^{t},x_{i}^{t+1}\right)\omega\left(g^{t+1}|x_{i}^{t+1}\right)\right]\end{array}\label{eq:rec-phi}\end{equation}

By using this identity, the maximum potential $\phi^{*}$ can be computed
efficiently from the partial maximum potentials $\phi_{i}^{t}$. The
computation of the trajectory that realizes this maximum potential
ensues by tracing back the computation to find which partial trajectory
realized $\phi_{i}^{t}$ for each step $t$.

\subsection{Trajectory filtering and smoothing}

\begin{figure}
\centering{}\includegraphics[width=6cm]{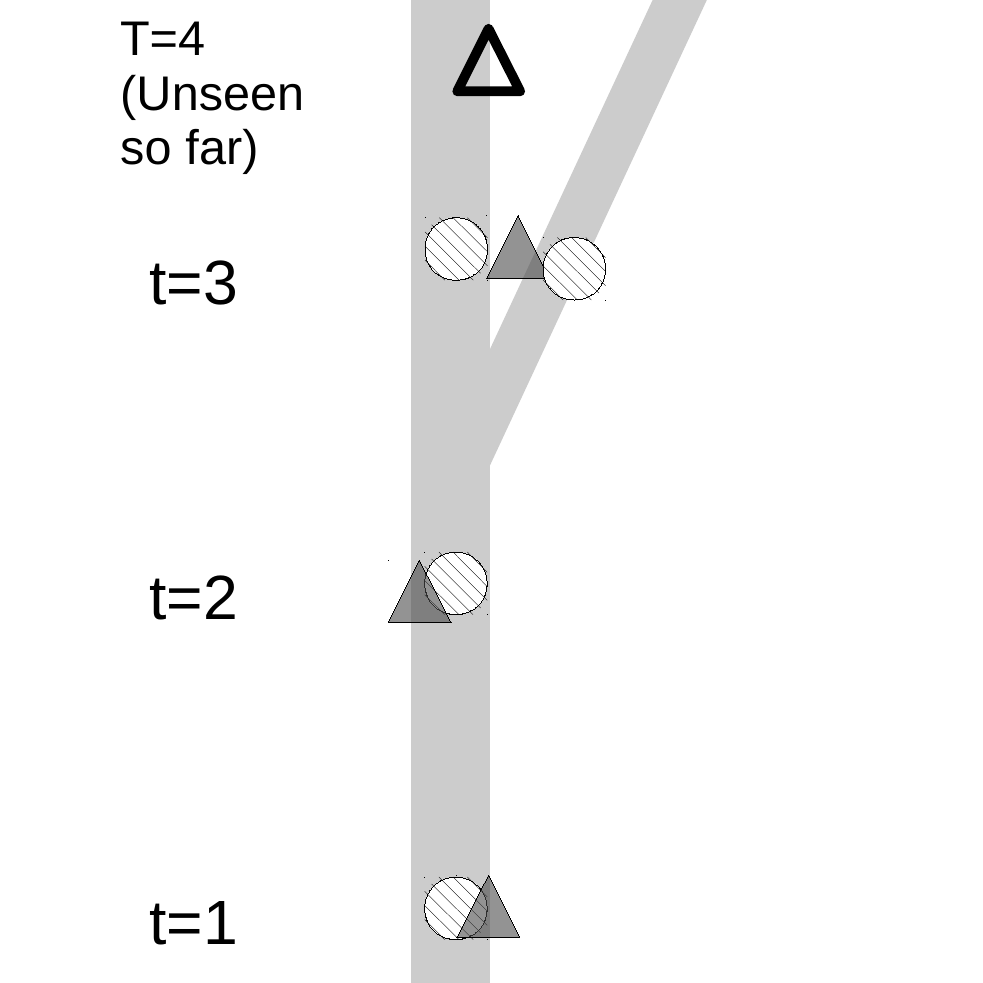}\caption{Example of case handled by lagged smoothing which disambiguates the
results provided by tracking. An observation is available close to
an exit ramp of a highway, for which the algorithm has to decide if
it corresponds to the vehicle exiting the highway. Lagged smoothing
analyzes subsequent points in the trajectory and can disambiguate
the situation.\label{fig:highway-exit}}
\end{figure}

\label{sub:filtering}We now turn our attention to the problem of
computing the marginals of the posterior distributions over the trajectories,
i.e. the probability distributions $\pi\left(x^{t}|g^{1:T}\right)$
and $\pi\left(p^{t}|g^{1:T}\right)$ for all $t$. We introduce some
additional notation to simplify the subsequent derivations. The posterior
probability $\bar{q}_{i}^{t}$ of the vehicle being at the state $x_{i}^{t}\in\mathbf{x}^{t}$
at time $t$, given all the observations $g^{1:T}$ of the trajectory,
is defined as:\[
\overline{q}_{i}^{t}\propto\pi\left(x_{i}^{t}|g^{1:T}\right)=\frac{1}{Z}\sum_{\tau=x^{1}\cdots p^{t-1}x_{i}^{t}p^{t}\cdots x^{T}}\phi\left(\tau|g^{1:T}\right)\]
The operator $\propto$ indicates that this distribution is defined
up to some scaling factor, which does not depend on $x$ or $p$ (but
may depend on $g^{1:T}$). Indeed, we are interested in the probabilistic
weight of a state $x_{i}^{t}$ relative to the other possible states
$x_{i'}^{t}$ at the state time $t$ (and not to the actual, unscaled
value of $\pi\left(x_{i}^{t}|g^{1:T}\right)$). This is why we consider
$\left(\bar{q}_{i}^{t}\right)_{i}$ as a choice between a (finite)
set of discrete variables, one choice per possible state $x_{i}^{t}$.
A natural choice is to scale the distribution $\bar{q}_{i}^{t}$ so
that the probabilistic weight of all possibilities is equal to 1:\[
\sum_{1\leq i\leq I^{t}}\bar{q}_{i}^{t}=1\]
From a practical perspective, $\bar{q}^{t}$ can be computed without
the knowledge of the partition function $Z$. Indeed, the only required
elements are the unscaled values of $\pi\left(x_{i}^{t}|g^{1:T}\right)$
for each $i$. The distribution $\bar{q}^{t}=\left(\bar{q}_{i}^{t}\right)_{i}$
is a multinomial distribution between $I^{t}$ choices, one for each
state. The quantity $\bar{q}_{i}^{t}$ has a clear meaning: it is
the probability that the vehicle is in state $x_{i}^{t}$, when choosing
amongst the set $\left(x_{i'}^{t}\right)_{1\leq i\leq I^{t}}$, given
all the observations $g^{1:T}$.

For each time $t$ and each path index $j\in\left[1\cdots J^{t}\right]$,
we also introduce (up to a scaling constant) the discrete distribution
over the paths at time $t$ given the observations $g^{1:T}$:

\[
\overline{r}_{j}^{t}\propto\pi\left(p_{j}^{t}|g^{1:T}\right)\]
which are scaled so that $\sum_{1\leq j\leq J^{t}}\overline{r}_{j}^{t}=1$.

This problem of smoothing in CRFs is a classic application of the
Junction Tree algorithm to chain-structured graphs \cite{murphy2002dynamic}.
For the sake of completeness, we derive an efficient smoothing algorithm
using our notations.

The definition of $\pi\left(x_{i}^{t}|g^{1:T}\right)$ requires summing
the potentials of all the trajectories that pass through the state
$x_{i}^{t}$ at time $t$. The key insight for efficient filtering
or smoothing is to make use of the chain structure of the graph, which
lets us factorize the summation into two terms, each of which can
be computed much faster than the exponentially large summation. Indeed,
one can show from the structure of the clique graph that the following
holds for all time steps $t$:\begin{equation}
\pi\left(x^{t}|g^{1:T}\right)\propto\pi\left(x^{t}|g^{1:t}\right)\pi\left(x^{t}|g^{t+1:T}\right)\label{eq:forback-decomposition}\end{equation}

The first term of the pair corresponds to the effect that the \emph{past
and present observations} ($g^{1:t}$) have on our belief of the present
state $x^{t}$. The second term corresponds to the effect that the
\emph{future observations }($g^{t+1:T}$) have on our estimation of
the present state. The terms $\pi\left(x^{t}|g^{1:t}\right)$ are
related to each other by an equation that propagates \emph{forward
}in time, while the terms $\pi\left(x^{t}|g^{t+1:T}\right)$ are related
through an equation that goes \emph{backward }in time. This is why
we call $\pi\left(x^{t}|g^{1:t}\right)$ the \emph{forward distribution}
\emph{for the states}, and we denote it %
\footnote{The arrow notation indicates that the computations for $\left.\overrightarrow{q}\right._{i}^{t}$
will be done forward in time.%
} by $\left(\left.\overrightarrow{q}\right._{i}^{t}\right)_{1\leq i\leq I^{t}}$:

\[
\left.\overrightarrow{q}\right._{i}^{t}\propto\pi\left(x_{i}^{t}|g^{1:t}\right)\]
The distribution $\left.\overrightarrow{q}\right._{i}^{t}$ is proportional
to the posterior probability $\pi\left(x_{i}^{t}|g^{1:t}\right)$
and the vector $\left.\overrightarrow{q}\right.^{t}=\left(\left.\overrightarrow{q}\right._{i}^{t}\right)_{i}$
is normalized so that $\sum_{i=1}^{I^{t}}\left.\overrightarrow{q}\right._{i}^{t}=1$.
We do this for the paths, by defining the \emph{forward distribution
for the paths}: \[
\left.\overrightarrow{r}\right._{j}^{t}\propto\pi\left(p_{j}^{t}|g^{1:t}\right)\]
Again, the distributions are defined up to a normalization factor
so that each component sums to $1$.

In the same fashion, we introduce the \emph{backward distributions
for the states and the paths:}

\[
\left.\overleftarrow{q}\right._{i}^{t}\propto\pi\left(x_{i}^{t}|g^{t+1:T}\right)\]
\[
\left.\overleftarrow{r}\right._{j}^{t}\propto\pi\left(p_{j}^{t}|g^{t+1:T}\right)\]
Using this set of notations, Equation \eqref{eq:forback-decomposition}
can be rewritten: \begin{eqnarray*}
\overline{q}_{i}^{t} & \propto & \left.\overrightarrow{q}\right._{i}^{t}\cdot\left.\overleftarrow{q}\right._{i}^{t}\\
\overline{r}_{j}^{t} & \propto & \left.\overrightarrow{r}\right._{j}^{t}\cdot\left.\overleftarrow{r}\right._{j}^{t}\end{eqnarray*}

Furthermore, $\left.\overrightarrow{r}\right.^{t}$ and $\left.\overrightarrow{q}\right.^{t}$
are related through a pair of recursive equations: \[
\left.\overrightarrow{q}\right._{i}^{1}\propto\pi\left(x_{i}^{1}|g^{1}\right)\]

\begin{equation}
\left.\overrightarrow{r}\right._{j}^{t}\propto\eta\left(p_{j}^{t}\right)\left(\sum_{j:\underline{\delta}\left(x_{i}^{t},p_{j}^{t}\right)=1}\left.\overrightarrow{q}\right._{i}^{t}\right)\label{eq:r_forward}\end{equation}
\begin{equation}
\left.\overrightarrow{q}\right._{i}^{t}\propto\omega\left(x_{i}^{t}|g^{t}\right)\left(\sum_{j:\bar{\delta}\left(p_{j}^{t-1},x_{i}^{t}\right)=1}\left.\overrightarrow{r}\right._{j}^{t-1}\right)\label{eq:q_forward}\end{equation}

Similarly, the backward distributions can be defined recursively,
starting from $t=T$: \[
\left.\overleftarrow{q}\right._{i}^{T}\propto1\]

\begin{equation}
\left.\overleftarrow{r}\right._{j}^{t}\propto\eta\left(p_{j}^{t}\right)\left(\sum_{j:\bar{\delta}\left(p_{j}^{t},x_{i}^{t+1}\right)=1}\left.\overleftarrow{q}\right._{i}^{t+1}\right)\label{eq:r_backward}\end{equation}

\begin{equation}
\left.\overleftarrow{q}\right._{i}^{t}\propto\omega\left(x_{i}^{t}|g^{t}\right)\left(\sum_{j:\underline{\delta}\left(x_{i}^{t},p_{j}^{t}\right)=1}\left.\overleftarrow{r}\right._{j}^{t}\right)\label{eq:q_backward}\end{equation}

Details of the forward algorithm and backward algorithm are provided
in the Algorithm \ref{alg:forward} and Algorithm \ref{alg:backward}
below. The complete algorithm for smoothing is detailed in the Algorithm
\ref{alg:smoothing} below.

\begin{algorithm}[h]
\begin{minipage}[t]{1\columnwidth}%
Given a sequence of observations $g^{1:T}$, a sequence of sets of
candidate projections $\mathbf{x}^{1:T}$ and a sequence of sets of
candidate paths $\mathbf{p}^{1:T-1}$:

Initialize the forward state distribution:

~~~~$\forall i=1\cdots I^{1}$: $\left.\overrightarrow{q}\right._{i}^{1}\leftarrow\omega\left(x_{i}^{1}|g^{1}\right)$

~~~~Normalize $\left.\overrightarrow{q}\right.^{1}$

For every time step $t$ from 1 to $T-1$:

~~~~Compute the forward probability over the paths:

~~~~$\begin{array}{lc}
\forall j=1\cdots J^{t}:\\
\,\,\,\left.\overrightarrow{r}\right._{j}^{t}\leftarrow\eta\left(p_{j}^{t}\right)\left(\sum_{j:\underline{\delta}\left(x_{i}^{t},p_{j}^{t}\right)=1}\left.\overrightarrow{q}\right._{i}^{t}\right)\end{array}$

~~~~Normalize $\left.\overrightarrow{r}\right.^{t}$

~~~~Compute the forward probability over the states:

~~~~$\begin{array}{lc}
\forall i=1\cdots I^{t+1}:\\
\,\,\,\left.\overrightarrow{q}\right._{i}^{t+1}\leftarrow\omega\left(x_{i}^{t+1}|g^{t+1}\right)\left(\sum_{j:\bar{\delta}\left(p_{j}^{t},x_{i}^{t+1}\right)=1}\left.\overrightarrow{r}\right._{j}^{t}\right)\end{array}$

~~~~Normalize $\left.\overrightarrow{q}\right.^{t+1}$

Return the set of vectors $\left(\left.\overrightarrow{q}\right.^{t}\right)_{t}$
and $\left(\left.\overrightarrow{r}\right.^{t}\right)_{t}$%
\end{minipage}

\caption{Description of forward recursion\label{alg:forward}}
\end{algorithm}

\begin{algorithm}[h]
\begin{minipage}[t]{1\columnwidth}%
Given a sequence of observations $g^{1:T}$, a sequence of sets of
candidate projections $\mathbf{x}^{1:T}$ and a sequence of sets of
candidate paths $\mathbf{p}^{1:T-1}$:

Initialize the backward state distribution

$\forall i=1\cdots I^{T}$: $\left.\overleftarrow{q}\right._{i}^{T}\leftarrow1$

For every time step $t$ from $T-1$ to 1:

~~~~Compute the forward probability over the paths:

~~~~$\forall j=1\cdots J^{t}:$

~~~~~~~$\left.\overleftarrow{r}\right._{j}^{t}\leftarrow\eta\left(p_{j}^{t}\right)\left(\sum_{j:\bar{\delta}\left(p_{j}^{t},x_{i}^{t+1}\right)=1}\left.\overleftarrow{q}\right._{i}^{t+1}\right)$

~~~~Normalize $\left.\overleftarrow{r}\right.^{t}$

~~~~Compute the forward probability over the states:

~~~~$\forall i=1\cdots I^{t}$:

~~~~~~~$\left.\overleftarrow{q}\right._{i}^{t}\leftarrow\omega\left(x_{i}^{t+1}|g^{t+1}\right)\left(\sum_{j:\underline{\delta}\left(x_{i}^{t},p_{j}^{t}\right)=1}\left.\overleftarrow{r}\right._{j}^{t}\right)$

~~~~Normalize $\left.\overleftarrow{q}\right.^{t}$

Return the set of vectors $\left(\left.\overleftarrow{q}\right.^{t}\right)_{t}$
and $\left(\left.\overleftarrow{r}\right.^{t}\right)_{t}$%
\end{minipage}

\caption{Description of backward recursion\label{alg:backward}}
\end{algorithm}

The above smoothing algorithm requires all the observations of a trajectory
in order to run. We have presented so far an \emph{a posteriori }algorithm
that requires full knowledge of measurements $g^{1:T}$. In this form,
it is not directly suitable for real-time applications that involve
streaming data, for which the data is available up to $t$ only. However,
this algorithm can be adapted for a variety of scenarios:
\begin{itemize}
\item \emph{Smoothing}, also called \emph{offline filtering}. This corresponds
to getting the best estimate given all observations, i.e. to computing
$\pi\left(x^{t}|g^{1:T}\right)$. The Algorithm \ref{alg:smoothing}
describes this procedure.
\item \emph{Tracking, filtering}, or \emph{online estimation.} This usage
corresponds to updating the current state of the vehicle as soon as
a new streaming observation is available, i.e. to computing $\pi\left(x^{t}|g^{1:t}\right)$.
This is exactly the case the forward algorithm (Algorithm \ref{alg:forward})
is set to solve. If one is simply interested in the most recent estimate,
then only the previous forward distribution $\left.\overrightarrow{q}\right.^{t}$
needs to be kept, and all distributions $\left.\overrightarrow{q}\right.^{t-1}\cdots\left.\overrightarrow{q}\right.^{1}$
at previous times can be discarded. This application minimizes the
latency and the computations at the expense of the accuracy.
\item \emph{Lagged smoothing}, or \emph{lagged filtering}. A few points
of data are stored and processed before returning a result. Algorithm
\ref{alg:lagged-smoothing} details this procedure, which involves
computing $\pi\left(x^{t}|g^{1:t+k}\right)$ for some $k>0$. A trade-off
is being made between the latency and the accuracy, as the information
from the points $g^{t+1:t+k}$ is used to update the estimate of the
state $x^{t}$. As shown in Section \ref{sec:results}, even for small
values of $k$, such a procedure can bring significant improvements
in the accuracy while keeping the latency within reasonable bounds.
A common ambiguity solved by lagged smoothing is presented in Figure
\ref{fig:highway-exit}.
\end{itemize}
\begin{algorithm}
\begin{minipage}[t]{1\columnwidth}%
Given a sequence of observations $g^{1:T}$, a sequence of sets of
candidate projections $\mathbf{x}^{1:T}$ and a sequence of sets of
candidate paths $\mathbf{p}^{1:T-1}$:

Compute $\left(\left.\overrightarrow{q}\right.^{t}\right)_{t}$ and
$\left(\left.\overrightarrow{r}\right.^{t}\right)_{t}$ using the
forward algorithm.

Compute $\left(\left.\overleftarrow{q}\right.^{t}\right)_{t}$ and
$\left(\left.\overleftarrow{r}\right.^{t}\right)_{t}$ using the backward
algorithm.

For every time step $t$:

~~~~$\forall j=1\cdots J^{t}:\overline{r}_{j}^{t}\leftarrow\left.\overrightarrow{r}\right._{j}^{t}\cdot\left.\overleftarrow{r}\right._{j}^{t}$

~~~~Normalize $\overline{r}^{t}$

~~~~$\forall i=1\cdots I^{t}$: $\overline{q}_{i}^{t}\leftarrow\left.\overrightarrow{q}\right._{i}^{t}\cdot\left.\overleftarrow{q}\right._{i}^{t}$

~~~~Normalize $\overline{q}^{t}$

Return the set of vectors $\left(\overline{q}^{t}\right)_{t}$ and
$\left(\overline{r}^{t}\right)_{t}$%
\end{minipage}

\caption{Trajectory smoothing algorithm\label{alg:smoothing}}
\end{algorithm}

\begin{algorithm}
\begin{minipage}[t]{1\columnwidth}%
Given an integer $k>0$, and a LIFO queue of observations:

Initialize the queue to the empty queue.

When receiving a new observation $g^{t}$:

~~~~Push the observation in the queue

~~~~Run the forward filter on this observation

~~~~If $t>k$:

~~~~~~~~Run the backward filter on the queue

~~~~~~~~Compute $\overline{q}^{t-k}$, $\overline{r}^{t-k}$
on the first element of the queue

~~~~~~~~Pop the queue and return $\overline{q}^{t-k}$ and
$\overline{r}^{t-k}$ %
\end{minipage}

\caption{Lagged smoothing algorithm\label{alg:lagged-smoothing}}
\end{algorithm}

\subsection{Observation model}

\label{sub:obs-mmodel}

We now describe the observation model $\omega$. The observation probability
is assumed only to depend on the distance between the point and the
GPS coordinates. We take an isoradial Gaussian noise model:\begin{eqnarray*}
\omega\left(g|x\right) & = & p\left(\text{d}\left(g,x\right)\right)\\
 & = & \frac{1}{\sqrt{2\pi}\sigma}\left(-\frac{1}{2\sigma^{2}}\text{d}\left(g,x\right)^{2}\right)\end{eqnarray*}
in which the function d is the distance function between geocoordinates.
The standard deviation $\sigma$ is assumed to be constant over all
the network. This is not true in practice because of well documented
urban canyoning effects \cite{cui2003autonomous,thiagarajan2010cooperative,thiagarajan2009vtrack}
and satellite occlusions. Updating the model accordingly presents
no fundamental difficulty, and can be done by geographical clustering
of the regions of interest. Using the estimation techniques described
later in Section \ref{sub:mle} and Section \ref{sub:em}, an estimate
of $\sigma$ between 10 and 15 meters could be estimated for data
of interest in this article.

\subsection{Driver model}

\label{sub:transition}

The second model to consider is the driver behavior model. This model
assigns a weight to any acceptable path on the road network. We consider
a model in the \emph{exponential family}, in which the weight distribution
over any path $p$ only depends on a selected number of features $\varphi\left(p\right)\in\mathbb{R}^{K}$
of the path. Possible features include the length of the path, the
number of stop signs, and the speed limits on the road. The distribution
is parametrized by a vector $\mu\in\mathbb{R}^{K}$ so that the logarithm
of the distribution of paths is a linear combination of the features
of the path:\[
\eta\left(p\right)\propto\exp\left(\mu^{T}\varphi\left(p\right)\right)\]
The function $\varphi$ is called the \emph{feature function, }and
the vector $\mu$ is called the behavioral \emph{parameter vector},
and simply encodes a weighted combination of the features.

In a simple model the vector $\varphi\left(p\right)$ may be reduced
to a single scalar, such as the length of the path. Then the inverse
of $\mu$, a length, can be interpreted as a characteristic length.
This model simply states that the driver has a preference for shorter
paths, and $\mu^{-1}$ indicates how aggressively this driver wants
to follow the shortest path. Such a model is explored in Section \eqref{sec:results}.
Other models considered include the mean speed and travel times, the
stop signs and signals, and the turns to the right or to the left.

In the \emph{Mobile Millennium} system, the path inference filter
is the input to a model designed to learn travel times, so the feature
function does not include dynamic features such as the current travel
time. Assuming this information is available, it would be easy to
add as a feature.

\section{Training procedure}

\label{sec:training}

The procedure detailed so far requires the calibration of the observation
model and the path selection model by setting some values for the
weight vector $\mu$ and the standard deviation $\sigma$. Using standard
machine learning techniques, we maximize the likelihood of the observations
with respect to the parameters, and we evaluate the result against
held-out trajectories using several metrics detailed in Section \ref{sec:results}.
Computing likelihood will require the computation of the partition
function (which depends on $\mu$ and $\sigma$). We first present
a procedure that is valid for any path or point distributions that
belong to the \emph{exponential family}, and then show how the models
we presented in Section \ref{sec:hmm} fit into this framework.

\subsection{Learning within the exponential family and sparse trajectories}

\label{sub:learning-exp-family}There is a striking similarity between
the state variables $x^{1:T}$ and the path variables $p^{1:T}$ \LyXbar{}
especially between the forward and backward distributions introduced
in Equation~\eqref{eq:r_forward}. This suggests to generalize our
procedure to a context larger than states interleaved with paths.
Indeed, each step of choosing a path or a variable corresponds to
making a \emph{choice }between a finite number of possibilities, and
there is a limited number of pairwise compatible choices (as encoded
by the functions $\underline{\delta}$ and $\bar{\delta}$). Following
a trajectory corresponds to choosing a new state (subject to the compatibility
constraints of the previous state). In this section, we introduce
the proper notation to generalize our learning problem, and then show
how this learning problem can be efficiently solved. In the next section,
we will describe the relation between the new variables we are going
to introduce and the parameters of our model.

Consider a joint sequence of multinomial random variables $\mathbf{z}^{1:L}=\mathbf{z}^{1}\cdots\mathbf{z}^{L}$
drawn from some space $\prod_{l=1}^{L}\left\{ 1\cdots K^{l}\right\} $
where $K^{l}$ is the dimensionality of the multinomial variable $\mathbf{z}^{l}$.
Given a realization $z^{1:L}$ from $\mathbf{z}^{1:L}$, we define
a non-negative potential function $\psi\left(z^{1:L}\right)$ over
the sequence of variables. This potential function is controlled by
a parameter vector $\theta\in\mathbb{R}^{M}$: $\psi\left(z^{1:L}\right)=\psi\left(z^{1:L};\theta\right)$
\footnote{The semicolon notation indicates that this function is parametrized
by $\theta$, but that $\theta$ is not a random variable.%
}. Furthermore, we assume that this potential function is also defined
and non-negative over any subsequence $\psi\left(z^{1:l}\right)$.
Lastly, we assume that there exists at least one sequence $z^{1:L}$
that has a positive potential. As in the previous section, the potential
function $\psi$, when properly normalized, defines a probability
distribution of density $\pi$ over the variables $\mathbf{z}$, and
this distribution is parametrized by the vector $\theta$:\begin{equation}
\pi\left(z;\theta\right)=\frac{\psi\left(z;\theta\right)}{Z\left(\theta\right)}\label{eq:distribution-generic}\end{equation}
with $Z=\sum_{z}\psi\left(z;\theta\right)$ called the \emph{partition
function}. We will show the partition function defined here is the
partition function introduced in Section \ref{sub:filtering}.

We assume that $\psi$ is an unscaled member of the \emph{exponential
family}: it is of the form:\begin{equation}
\psi\left(z;\theta\right)=h\left(z\right)\prod_{l=1}^{L}e^{\theta\cdot T^{l}\left(z^{l}\right)}\label{eq:expo-family}\end{equation}
In this representation, $h$ is a non-negative function of $z$ which
does not depend on the parameters, the operator $\cdot$ is the vector
dot product, and the vectors $T^{l}\left(z^{l}\right)$ are vector
mappings from the realization $z^{l}$ to $\mathbb{R}^{M}$ for some
$M\in\mathbb{N}$, called \emph{feature vectors}. Since the variable
$z^{l}$ is discrete and takes on values in $\left\{ 1\cdots K^{l}\right\} $,
it is convenient to have a specific notation for the feature vector
associated with each value of this variable:\[
\forall i\in\left\{ 1\cdots K^{l}\right\} ,T_{i}^{l}=T^{l}\left(z^{l}=i\right)\]

The sequence of variables \textbf{$\mathbf{z}$ }represents the choices
associated with a single trajectory, i.e. the concatenation of the
$x$s and $p$s. In general, we will observe and would like to learn
from multiple trajectories at the same time. This is why we need to
consider a collection of variables $\left(\mathbf{z}^{\left(u\right)}\right)_{u}$,
each of which follows the form above and each of which we can define
a potential $\psi\left(z^{\left(u\right)};\theta\right)$ and a partition
function $Z^{\left(u\right)}\left(\theta\right)$ for. There the variable
$u$ indexes the set of sequences of observations, i.e. the set of
consecutive GPS measurements of a vehicle. Since each of these trajectories
will take place on a different portion of the road network, each of
the sequences $\mathbf{z}^{\left(u\right)}$ will have a different
state space. For each of these sequences of variables $\mathbf{z}^{\left(u\right)}$,
we observe the respective realizations $z^{\left(u\right)}$ (which
correspond to the observation of a trajectory), and we wish to infer
the parameter vector $\theta^{*}$ that maximizes the likelihood of
all the realizations of the trajectories:\begin{equation}
\begin{aligned}\theta^{*} & =\underset{\theta}{\arg\max}\sum_{u}\log\pi^{\left(u\right)}\left(z^{\left(u\right)};\theta\right)\\
 & =\underset{\theta}{\arg\max}\sum_{u}\log\psi\left(z^{\left(u\right)};\theta\right)-\log Z^{\left(u\right)}\left(\theta\right)\\
 & =\underset{\theta}{\arg\max}\sum_{u}\sum_{l=1}^{L^{\left(u\right)}}\theta\cdot T^{l^{\left(u\right)}}\left(z^{l^{\left(u\right)}}\right)-\log Z^{\left(u\right)}\left(\theta\right)\end{aligned}
\label{eq:MaxLL-problem}\end{equation}
where again the indexing $u$ is for sets of measurements of a given
trajectory. Similarly, the length of a trajectory is indexed by $u$:
$L^{\left(u\right)}$. From Equation \ref{eq:MaxLL-problem}, it is
clear that the log-likelihood function simply sums together the respective
likelihood functions of each trajectory. For clarity, we consider
a single sequence $z^{\left(u\right)}$ only and we remove the indexing
with respect to $u$. With this simplification, we have for a single
trajectory: \begin{equation}
\begin{aligned}\log\psi\left(z;\theta\right)-\log Z\left(\theta\right) & =\sum_{l=1}^{L}\theta\cdot T^{l}\left(z^{l}\right)-\log Z\left(\theta\right)\end{aligned}
\label{eq:z-exp}\end{equation}

The first part of Equation \eqref{eq:z-exp} is linear with respect
to $\theta$ and $\log Z\left(\theta\right)$ is concave in $\theta$
(it is the logarithm of a sum of exponentiated linear combinations
of $\theta$ \cite{boyd2004convex}) . As such, maximizing Equation
\eqref{eq:z-exp} yields a unique solution (assuming no singular parametrization),
and some superlinear algorithms exist to solve this equation \cite{boyd2004convex}.
These algorithms rely on the computation of the gradient and the Hessian
matrix of $\log Z\left(\theta\right)$. We now detail some closed-form
recursive formulas to compute these elements.

\subsubsection{Efficient estimation of the partition function}

A naive approach to the computation of the partition function $Z\left(\theta\right)$
leads to consider %
{}exponentially many paths. Most of these computations can be factored
using dynamic programming %
\footnote{This is - again - a specific application of the junction tree algorithm.
See \cite{murphy2002dynamic} for an explanation of the general framework.%
}. Recall the definition of the partition function:\[
Z\left(\theta\right)=\sum_{z}h\left(z\right)\prod_{l=1}^{L}e^{\theta\cdot T^{l}\left(z^{l}\right)}\]
So far, the function $h$ was defined in in a generic way (it is non-negative
and does not depend on $\theta$). We consider a particular shape
that generalizes the functions $\underline{\delta}$ and $\bar{\delta}$
introduced in the previous section. In particular, the function $h$
is assumed to be a binary function, from the Cartesian space $\prod_{l=1}^{L}\left\{ 1\cdots K^{l}\right\} $
to $\left\{ 0,1\right\} $, that decomposes to the product of binary
functions over consecutive pairs of variables:\[
h\left(z\right)=\prod_{l=1}^{L-1}h^{l}\left(z^{l},z^{l-1}\right)\]
in which every function $h^{l}$ is a binary indicator $h^{l}:\left\{ 1\cdots K^{l}\right\} \times\left\{ 1\cdots K^{l-1}\right\} \rightarrow\left\{ 0,1\right\} $.
These functions $h^{l}$ generalize the functions $\underline{\delta}$
and $\overline{\delta}$ for arguments $z$ equal to either the $x$s
or the $p$s. It indicates the compatibility of the values of the
instantiations $z^{l}$ and $z^{l-1}$ 

Finally, we introduce the following notation. For each index $l\in\left[1\cdots L\right]$
and subindex $i\in[1\cdots K^{l}]$, we call $Z_{i}^{l}$ the partial
summation of all partial paths $\mathbf{z}{}^{1:l}$ that terminate
at the value $z^{l}=i$:\begin{eqnarray*}
Z_{i}^{l}\left(\theta\right) & = & \sum_{z^{1:l}:z^{l}=i}h\left(z^{1:l}\right)\prod_{m=1}^{l}e^{\theta\cdot T^{m}\left(z^{m}\right)}\\
 & = & \sum_{z^{1:l}:z^{l}=i}e^{\theta\cdot T^{1}\left(z^{1}\right)}\prod_{m=2}^{l}h^{m}\left(z^{m},z^{m-1}\right)e^{\theta\cdot T^{m}\left(z^{m}\right)}\end{eqnarray*}
This partial summation can also be defined recursively:\begin{equation}
\begin{aligned}Z_{i}^{l}\left(\theta\right) & =e^{\theta\cdot T_{i}^{l}}\sum_{j\in\left[1...K^{l-1}\right]:h^{l}\left(z^{i},z^{j}\right)=1}Z_{j}^{l-1}\left(\theta\right)\end{aligned}
\label{eq:Z-recursive-definition}\end{equation}
The start of the recursion is for all $i\in\left\{ 1\cdots K^{1}\right\} $:\[
Z_{i}^{1}\left(\theta\right)=e^{\theta\cdot T_{i}^{1}}\]
 and the complete partition function is the summation of the auxiliary
values:\[
Z\left(\theta\right)=\sum_{i=1}^{K^{L}}Z_{i}^{L}\left(\theta\right)\]

Computing the partition function can be done in polynomial time complexity
by a simple application of dynamic programming. By using sparse data
structures to implement $h$, some additional savings in computations
can be made%
\footnote{In particular, care should be taken to implement all the relevant
computations in log-domain due to the limited precision of floating
point arithmetic on computers. The reference implementation \cite{pythonimpl}
shows one way to do it.%
}.

\subsubsection{Estimation of the gradient}

The estimation of the gradient for the first part of the log likelihood
function is straightforward. The gradient of the partition function
can also be computed using Equation \eqref{eq:Z-recursive-definition}:\[
\begin{aligned}\nabla_{\theta}Z_{i}^{l}\left(\theta\right) & =Z_{i}^{l}\left(\theta\right)T_{i}^{l}+e^{\theta\cdot T_{i}^{l}}\sum_{j:h^{l}\left(z^{i},z^{j}\right)=1}\nabla_{\theta}Z_{j}^{l-1}\left(\theta\right)\end{aligned}
\]
The Hessian matrix can be evaluated in similar fashion:

\[
\begin{aligned}\triangle_{\theta\theta}Z_{i}^{l}\left(\theta\right)= & Z_{i}^{l}\left(\theta\right)\left(T_{i}^{l}\right)\left(T_{i}^{l}\right)^{'}\\
 & +e^{\theta\cdot T_{i}^{l}}\left(\sum_{j:h^{l}\left(z^{i},z^{j}\right)=1}\nabla_{\theta}Z_{j}^{l-1}\left(\theta\right)\right)\left(T_{i}^{l}\right)^{'}\\
 & +e^{\theta\cdot T_{i}^{l}}\left(T_{i}^{l}\right)\left(\sum_{j:h^{l}\left(z^{i},z^{j}\right)=1}\nabla_{\theta}Z_{j}^{l-1}\left(\theta\right)\right)^{'}\\
 & +e^{\theta\cdot T_{i}^{l}}\sum_{j:h^{l}\left(z^{i},z^{j}\right)=1}\triangle_{\theta\theta}Z_{j}^{l-1}\left(\theta\right)\end{aligned}
\]

\subsection{Exponential family models}

We now express our formulation of Conditional Random Fields to a form
compatible with Equation~\eqref{eq:expo-family}.

Consider $\epsilon=\sigma^{-2}$ and $\theta$ the stacked vector
of the desired parameters: \[
\theta=\left(\begin{array}{c}
\epsilon\\
\mu\end{array}\right)\]

There is a direct correspondence between the path and state variables
with the $\mathbf{z}$ variables introduced above. Let us pose $L=2T-1$,
then for all $l\in\left[1,L\right]$ we have:\[
z^{2t}=r^{t}\]
\[
z^{2t-1}=q^{t}\]

and the feature vectors are simply the alternating values of $\varphi$
and d, completed by some zero values:\[
T_{i}^{2t}=\left(\begin{array}{c}
0\\
\varphi\left(p_{i}^{t}\right)\end{array}\right)\]
\[
T_{j}^{2t-1}=\left(\begin{array}{c}
-\frac{1}{2}\text{d}\left(g,x_{j}^{t}\right)^{2}\\
\mathbf{0}\end{array}\right)\]
These formulas establish how we can transform our learning problem
that involves paths and states into a more abstract problem that considers
a single set of variables.

\subsection{Supervised learning with known trajectories}

\label{sub:mle}The most straightforward way to learn $\mu$ and $\sigma$,
or equivalently to learn the joint vector $\theta$, is to maximize
the likelihood of some GPS observations $g^{1:T}$, knowing the complete
trajectory followed by the vehicle. For all time $t$, we also know
which path $p_{\text{observed}}^{t}$ was taken and which state $x_{\text{observed}}^{t}$
produced the GPS observation $g^{t}$. We make the assumption that
the observed path $p_{\text{observed}}^{t}$ is one of the possible
path amongst the set of candidate paths $\left(p_{j}^{t}\right)_{j}$:
\[
\exists j\in\left[1\cdots J^{t}\right]\;:\; p_{\text{observed}}^{t}=p_{j}^{t}\]
and similarly, that the observed state $x_{\text{observed}}^{t}$
is one of the possible states:\[
\exists i\in\left[1\cdots I^{t}\right]\;:\; x_{\text{observed}}^{t}=x_{i}^{t}\]
In this case, the values of $r^{t}$ and $q^{t}$ are known (they
are the matching indexes), and the optimization problem of Equation\,\eqref{eq:MaxLL-problem}
can be solved using methods outlined in Section~\ref{sub:learning-exp-family}.

\subsection{Unsupervised learning with incomplete observations: Expectation-Maximization}

\label{sub:em}

Usually, only the GPS observations $g^{1:T}$ are available; the values
of $r^{1:T-1}$ and $q^{1:T}$ (and thus $z^{1:L}$) are hidden to
us. In this case, we estimate the \emph{expected likelihood $\mathcal{L}$},
which is the expected value of the likelihood under the distribution
over the assignment variables \textbf{$\mathbf{z}^{1:L}$}:

\textbf{\begin{eqnarray}
\mathcal{L}\left(\theta\right) & = & \mathbb{E}_{z\sim\pi\left(\cdot|\theta\right)}\left[\log\left(\pi\left(z;\theta\right)\right)\right]\label{eq:MaxLL-expected}\\
 & = & \sum_{z}\pi\left(z;\theta\right)\log\left(\pi\left(z;\theta\right)\right)\end{eqnarray}
}The intuition behind this expression is quite natural: since we do
not know the value of the assignment variable $z$, we consider the
\emph{expectation} of the likelihood over this variable. This expectation
is done with respect to the distribution $\pi\left(z;\theta\right)$.
The challenge lies in the dependency in $\theta$ of the very distribution
used to take the expectation. Computing the expected likelihood becomes
much more complicated than simply solving the optimization problem
of \eqref{eq:MaxLL-problem}. 

One strategy is to find some {}``fill in'' values for $z$ that
would correspond to our guesses of which path was taken, and which
point made the observation. However, such a guess would likely involve
our model for the data, which we are currently trying to learn. A
solution to this chicken and egg problem is the Expectation Maximization
(EM) algorithm \cite{moon1996expectation}. This algorithm performs
an iterative projection ascent by assigning some \emph{distributions}
(rather than singular values) to every $z^{l}$, and uses these distributions
to updates the parameters $\mu$ and $\sigma$ using the procedures
seen in Section \ref{sub:mle}. This iterative procedure performs
two steps:
\begin{enumerate}
\item Fixing some value for $\theta$, it computes a distribution $\tilde{\pi}\left(z\right)=\pi\left(z;\theta\right)$
\item It then uses this distribution $\tilde{\pi}\left(z\right)$ to compute
some new value of $\theta$ by solving the approximate problem in
which the expectation is fixed with respect to $\theta$:\begin{equation}
\max_{\theta}\mathbb{E}_{z\sim\tilde{\pi}\left(\cdot\right)}\left[\log\left(\pi\left(z;\theta\right)\right)\right]\label{eq:MaxLL-EM}\end{equation}
This problem is significantly simpler than the optimization problem
in Equation \eqref{eq:MaxLL-expected} since the expectation itself
does not depend on $\theta$ and thus is not part of the optimization
problem.
\end{enumerate}
Under this procedure, the expected likelihood is shown to converge
to a local maximum \cite{murphy2002dynamic}. It can be shown that
good values for the plug-in distribution $\tilde{\pi}$ are simply
the values of the posterior distributions $\pi\left(p^{t}|g^{1:T}\right)$
and $\pi\left(x^{t}|g^{1:T}\right)$, i.e. the values $\overline{q}^{t}$
and $\bar{r}^{t}$. Furthermore, owing to the particular shape of
the distribution $\pi\left(z\right)$, taking the expectation is a
simple task: we simply replace the value of the feature vector by
its \emph{expected value }under the distribution $\tilde{\pi}\left(z\right)$.
More practically, we simply have to consider:\begin{equation}
\begin{aligned}T^{2t}\left(z^{2t}\right) & =\mathbb{E}_{p\sim\pi\left(\cdot|\theta,g^{1:T}\right)}\left[\left(\begin{array}{c}
0\\
\varphi\left(p_{r}^{t}\right)\end{array}\right)\right]\\
 & =\left(\begin{array}{c}
0\\
\mathbb{E}_{p\sim\pi\left(\theta,g^{1:T}\right)}\left[\varphi\left(p_{r}^{t}\right)\right]\end{array}\right)\end{aligned}
\label{eq:T-expected-paths}\end{equation}
in which \[
\mathbb{E}_{p\sim\pi\left(\theta,g^{1:T}\right)}\left[\varphi\left(p_{r}^{t}\right)\right]=\sum_{i=1}^{I^{t}}\overline{r}_{i}^{t}\varphi_{i}^{t}\]

and \begin{equation}
T^{2t-1}\left(z^{2t-1}\right)=\left(\begin{array}{c}
-\frac{1}{2}\mathbb{E}_{x\sim\pi\left(\cdot|\theta,g^{1:T}\right)}\left[\text{d}\left(g,x_{q^{t}}^{t}\right)^{2}\right]\\
\mathbf{0}\end{array}\right)\label{eq:T-expected-states}\end{equation}
so that\[
\mathbb{E}_{x\sim\pi\left(\cdot|\theta,g^{1:T}\right)}\left[\text{d}\left(g,x_{q^{t}}^{t}\right)^{2}\right]=\sum_{i=1}^{J^{t}}\bar{q}_{i}^{t}\text{d}\left(g,x_{i}^{t}\right)^{2}\]
These values of feature vectors plug directly into the supervised
learning problem in Equation \eqref{eq:MaxLL-problem} and produce
updated parameters $\mu$ and $\sigma$, which are then used in turn
for updating the values of $\bar{q}$ and $\bar{r}$ and so on.

\begin{algorithm}[t]
\begin{minipage}[t]{1\columnwidth}%
Given a set of sequences of observations, an initial value of $\theta$

Repeat until convergence:

~~~~For each sequence, compute $\overline{r}^{t}$ and $\overline{q}^{t}$
using Algorithm \ref{alg:smoothing}.

~~~~For each sequence, update expected values of $T^{t}$ using
\eqref{eq:T-expected-paths} and \eqref{eq:T-expected-states}.

~~~~Compute a solution of Problem \eqref{eq:MaxLL-problem} using
these new values of $T^{t}$.%
\end{minipage}

\caption{Expectation maximization algorithm for learning parameters without
complete observations.\label{Flo:em}}
\end{algorithm}

\section{Results from field operational test}

The path inference filter and its learning procedures were tested
using field data through the \emph{Mobile Millennium} system. Ten
San Francisco taxicabs were fit with high frequency GPS (1 second
sampling rate) in October 2010 during a two-day experiment. Together,
they collected about seven hundred thousand measurement points that
provided a high-accuracy ground truth. Additionally, the unsupervised
learning filtering was tested on a significantly larger dataset: one
day one-minute samples of 600 taxis from the same fleet, which represents
600 000 points. For technical reasons, the two datasets could not
be collected the same day, but were collected the same day of the
week (a Wednesday) three weeks prior to the high-frequency collection
campaign. Even if the GPS equipment was different, both datasets presented
the same distribution of GPS dispersion. Thus we evaluate two datasets
collected from the same source with the same spatial features: a smaller
set at high frequency, called {}``Dataset 1'', and a larger dataset
sampled at 1 minute for which we do not know ground truth, called
{}``Dataset 2''.

\subsection{Experiment design}

\begin{figure}
\begin{centering}
\includegraphics[width=0.5\textwidth]{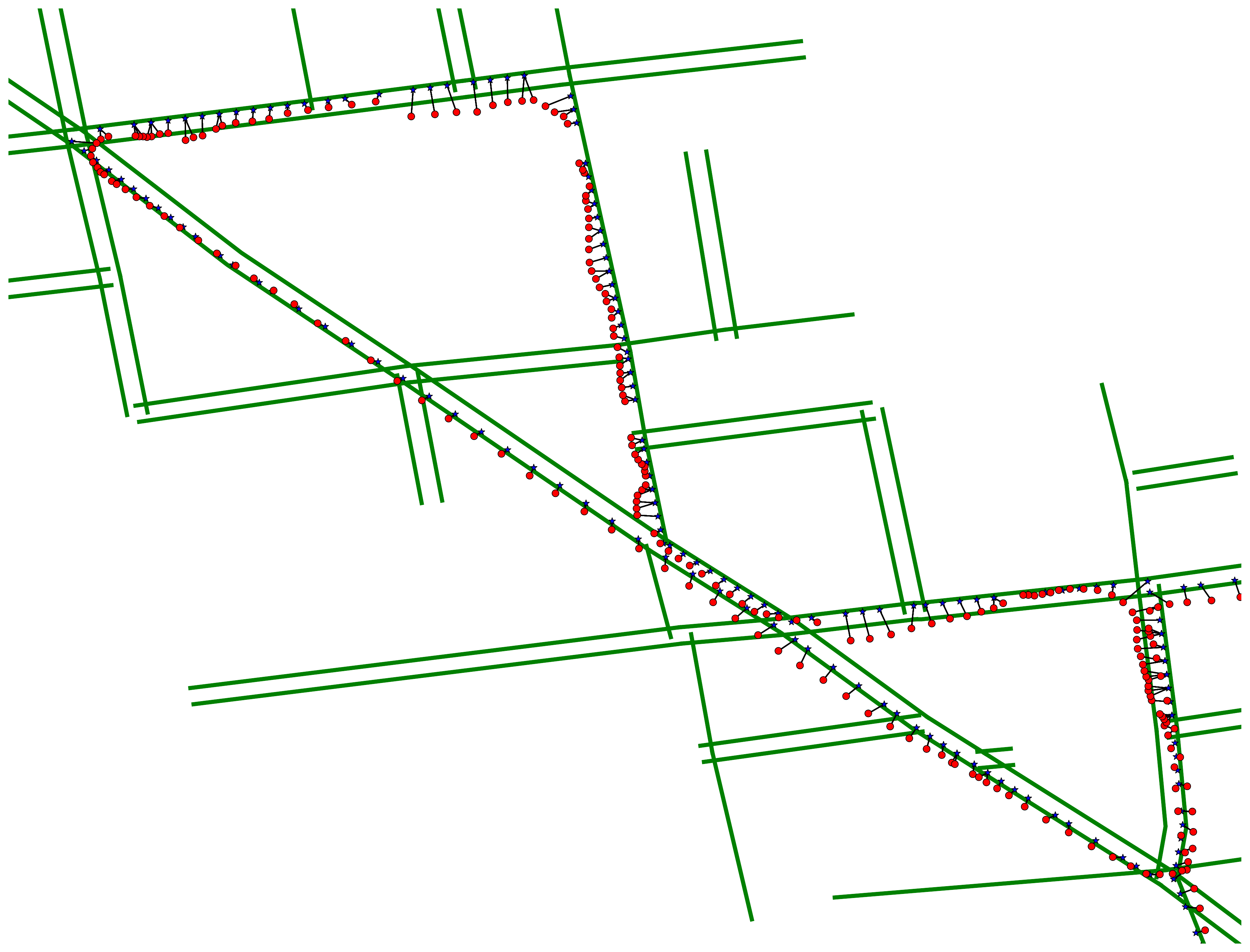}
\par\end{centering}

\caption{Example of points collected in {}``Dataset 1'', in the Russian Hill
neighborhood in San Francisco. The (red) dots are the GPS observations
(collected every second), and the green lines are road links that
contain a state projection. The black lines show the most likely projection
of the GPS points on the road network, using the Viterbi algorithm
on a gridded state-space with a 1-meter grid for the offsets.}
\end{figure}

\begin{algorithm}
Given a set of high-frequency sequences of raw GPS data:
\begin{enumerate}
\item Map the raw high-frequency sequences on the road network
\item Run the Viterbi algorithm with default settings
\item Extract the most likely HF trajectory on the road network for each
sequence
\item Given a set of projected HF trajectories:

\begin{enumerate}
\item Decimate the trajectories to a given sampling rate
\item Separate the set into a training subset and a test subset
\item Compute the best model parameters for a number of learning methods
(most likely, EM with a simple model or a more complex model)
\item Evaluate the model parameters with respect to different computing
strategies (Viterbi, online, offline, lagged smoothing) on the test
subset
\end{enumerate}
\end{enumerate}
\caption{Evaluation procedure}
\label{enu:evaluation-procedure}%
\end{algorithm}
The testing procedure is described in Algorithm \ref{enu:evaluation-procedure}:
the filter was first run in trajectory reconstruction mode (Viterbi
algorithm) with settings and-tuned for a high-frequency application,
using all the samples, in order to build a set of ground truth trajectories.
The trajectories were then downsampled to different temporal resolutions
and were used to test the filter in different configurations. The
following features were tested:
\begin{itemize}
\item The sampling rate. The following values were tested: 1 second, 10
seconds, 30 seconds, one minute, one and a half minute and two minutes
\item The computing strategy: pure filtering ({}``online'' or forward
filtering), fixed-lagged smoothing with a one- or two-point buffer
({}``1-lag'' and {}``2-lag'' strategies), Viterbi and smoothing
({}``offline'', or forward-backward procedure).
\item Different models:

\begin{itemize}
\item {}``Hard closest point'': A greedy deterministic model that computes
the closest point and then finds the shortest path to reach this closest
point from the previous point. This non-probabilistic model is the
baseline against which we make comparison on \cite{greenfeld2002matching}.
This greedy model may lead to non-feasible trajectories, for example
by assigning an observation to a dead end link from which it cannot
recover.
\item {}``Closest point'' : A non-greedy version of {}``Hard closest
point''. Among all the feasible trajectories, this (naive, deterministic)
model projects all the GPS data to their closest projections and then
selects the shortest path between each projection. The computing strategy
chosen is important because the filter may determine that some projections
lead to dead end and force the trajectory to break.
\item {}``Shortest path'': A naive model that selects the shortest path.
Given paths of the same length, it will take the path leading to the
closest point. The points projections are then recovered from the
paths. This is similar to \cite{giovannininovel,white2000some}.
\item {}``Simple'' A simple model that considers two features that could
be tuned by hand:

\begin{enumerate}
\item $\xi_{1}$ : The length of the path
\item $\xi_{2}$ : The distance of a point projection to its GPS coordinate
\end{enumerate}
This model was trained on learning data by two procedures: 
\begin{itemize}
\item Supervised learning, in which the true trajectory is provided to the
learning algorithm leading to the {}``MaxLL-Simple'' model
\item Unsupervised learning, which produced the model called {}``EM-Simple''
\end{itemize}
\item {}``Complex'' : A more complex model with a more diverse set of
features, which is complicated enough to discourage manual tuning:

\begin{enumerate}
\item The length of the path
\item The number of stop signs along the path
\item The number of signals (red lights)
\item The number of left turns made by the vehicle at road intersections
\item The number of right turns made by the vehicle at road intersections
\item The minimum average travel time (based on the speed limit)
\item The maximum average speed
\item The maximum number of lanes (representative of the class of the road)
\item The minimum number of lanes
\item The distance of a point to its GPS point
\end{enumerate}
This model was first evaluated using supervised learning leading to
the model called {}``MaxLL-Complex''. The unsupervised learning
procedure was also tried but failed to properly converge when using
{}``Dataset 1'', obtained from high-frequency samples. Unsupervised
learning was run again with {}``Dataset 2'', using the simple model
as a start point and converged properly this time. This set of parameters
is presented under the label {}``EM-Complex''.

\end{itemize}
\end{itemize}
All the models above are specific cases of our framework: 
\begin{itemize}
\item {}``Simple'' is a specific case of {}``Complex'', by restricting
the complex model to only two features.
\item {}``Shortest path'' is a specific case of {}``Simple'' with $\left|\xi_{1}\right|\gg1$,
$\left|\xi_{2}\right|\ll1$. We used $\xi_{1}=-1000$ and $\xi_{2}=-0.001$
\item {}``Closest point'' is a specific case of {}``Simple'' with $\left|\xi_{1}\right|\ll1$,
$\left|\xi_{2}\right|\gg1$. We used $\xi_{1}=-0.001$ and $\xi_{2}=-1000$
\item {}``Hard closest point'' can be reasonably approximated by running
the {}``Closest point'' model with the Online filtering strategy.
\end{itemize}
Thanks to this observation, we implemented all the model using the
same code and simply changed the set of features and the parameters
\cite{pythonimpl}.

These models were evaluated under a number of metrics:
\begin{itemize}
\item The proportion of path misses: for each trajectory, it is the number
of times the most likely path was not the true path followed, divided
by the number of time steps in the trajectory.
\item The proportion of state misses: for each trajectory, the number of
times the most likely projection was not the true projection. 
\item The log-likelihood of the true point projection. This is indicative
of how often the true point is identified by the model.
\item The log-likelihood of the true path.
\item The entropy of the path distribution and of the point distribution.
This statistical measure indicates the confidence assigned by the
filter to its result. A small entropy (close to 0) indicates that
one path is strongly favored by the filter against all the other ones,
whereas a large entropy indicates that all paths are equal.
\item The miscoverage of the route. Given two paths $p$ and $p'$ the coverage
of $p$ by $p'$, denoted $\text{cov}\left(p,p'\right)$ is the amount
of length of $p$ that is shared with $p'$ (it is a semi-distance
since it is not symmetric). It is thus lower than the total length
$\left|p\right|$ of the path $p$. We measure the dissimilarity of
two paths by the \emph{relative miscoverage}: $\text{mc}\left(p\right)=1-\frac{\text{cov}\left(p^{*},p\right)}{\left|p^{*}\right|}$.
If a path is perfectly covered, its relative miscoverage will be 0. 
\end{itemize}
For about 0.06\% of pairs of points, the true path could not be found
by the A{*} algorithm and was manually added to the set of discovered
paths

Each training session was evaluated with k-fold cross-validation,
using the following parameters:

\begin{center}
\begin{tabular}{|>{\centering}p{1.5cm}|>{\centering}p{1.5cm}|>{\centering}p{1.5cm}|}
\hline 
Sampling rate (seconds) & Batches used for validation & Batches used for training\tabularnewline
\hline
\hline 
1 & 1 & 5\tabularnewline
\hline 
10 & 3 & 5\tabularnewline
\hline 
30 & 6 & 5\tabularnewline
\hline 
60 & 6 & 5\tabularnewline
\hline 
90 & 6 & 5\tabularnewline
\hline 
120 & 6 & 5\tabularnewline
\hline
\end{tabular}
\par\end{center}

\subsection{Results}

Given the number of parameters to adjust, we only present the most
salient results here.

\begin{figure}
\begin{centering}
\includegraphics[width=3.5in]{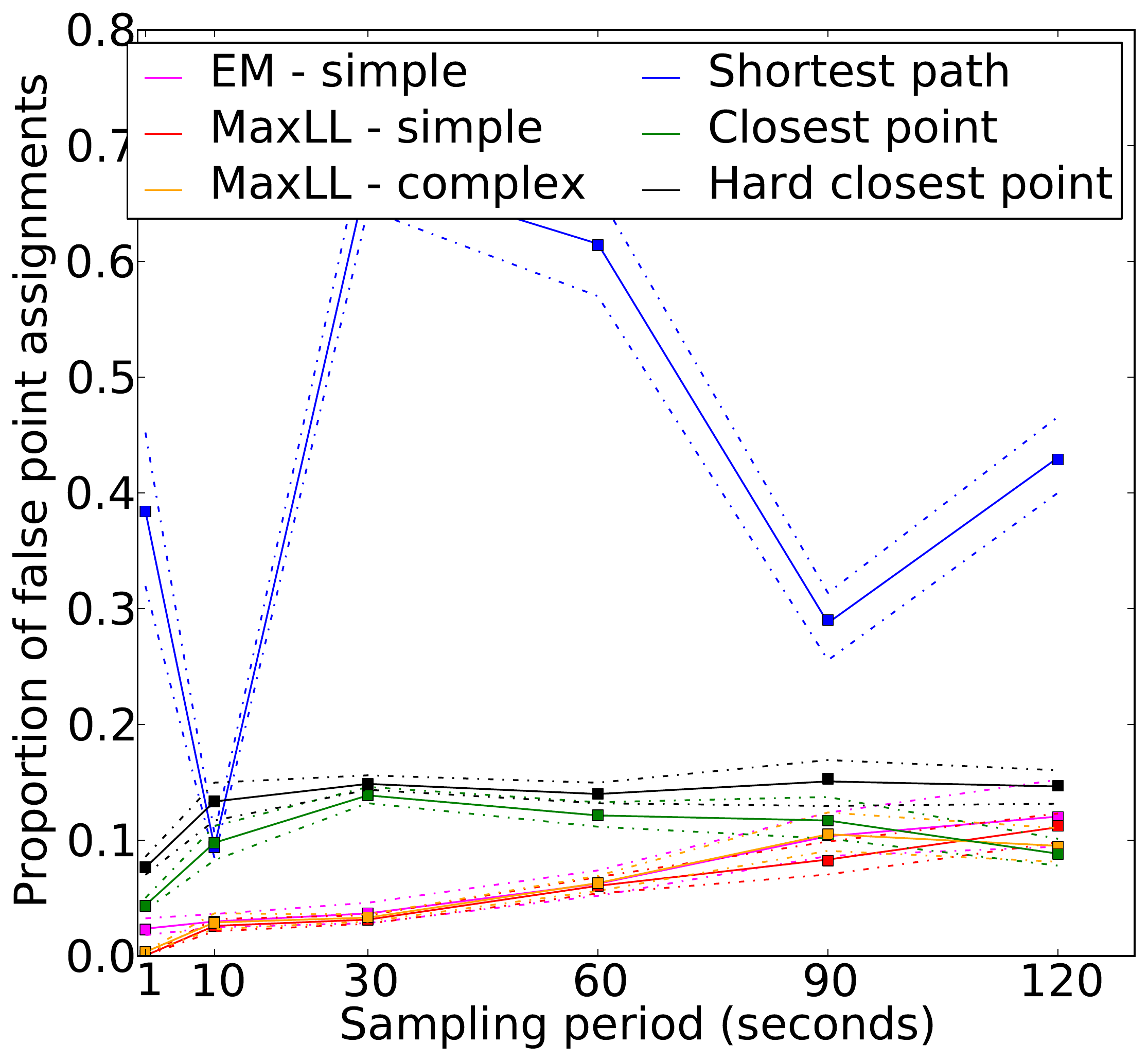}
\par\end{centering}

\caption{\label{Flo:point_miss}Point misses using trajectory reconstruction
(Viterbi algorithm) for different sampling rates, as a percentage
of incorrect point reconstructions for each trajectory (positive,
smaller is better). The solid line denotes the median, the squares
denote the mean and the dashed lines denote the 94\% confidence interval.
The black curve is the performance of a greedy reconstruction algorithm,
and the colored plots are the performances of probabilistic algorithms
for different features and weights learned by different methods. As
expected, the error rate is close to 0 for high frequencies (low sampling
rates): all the points are correctly identified by all the algorithms.
In the low frequencies (high sampling rates), the error still stays
low (around 10\%) for the probabilistic models, and also for the greedy
model. For sampling rates between 10 seconds and 90 seconds, tuned
models show a much higher performance compared to greedy models (Hard
closest point, closest point and shortest path). However, we will
see that the errors made by tuned models are more benign than errors
made by simple greedy models.}
\end{figure}

\begin{figure}
\begin{centering}
\includegraphics[width=3in]{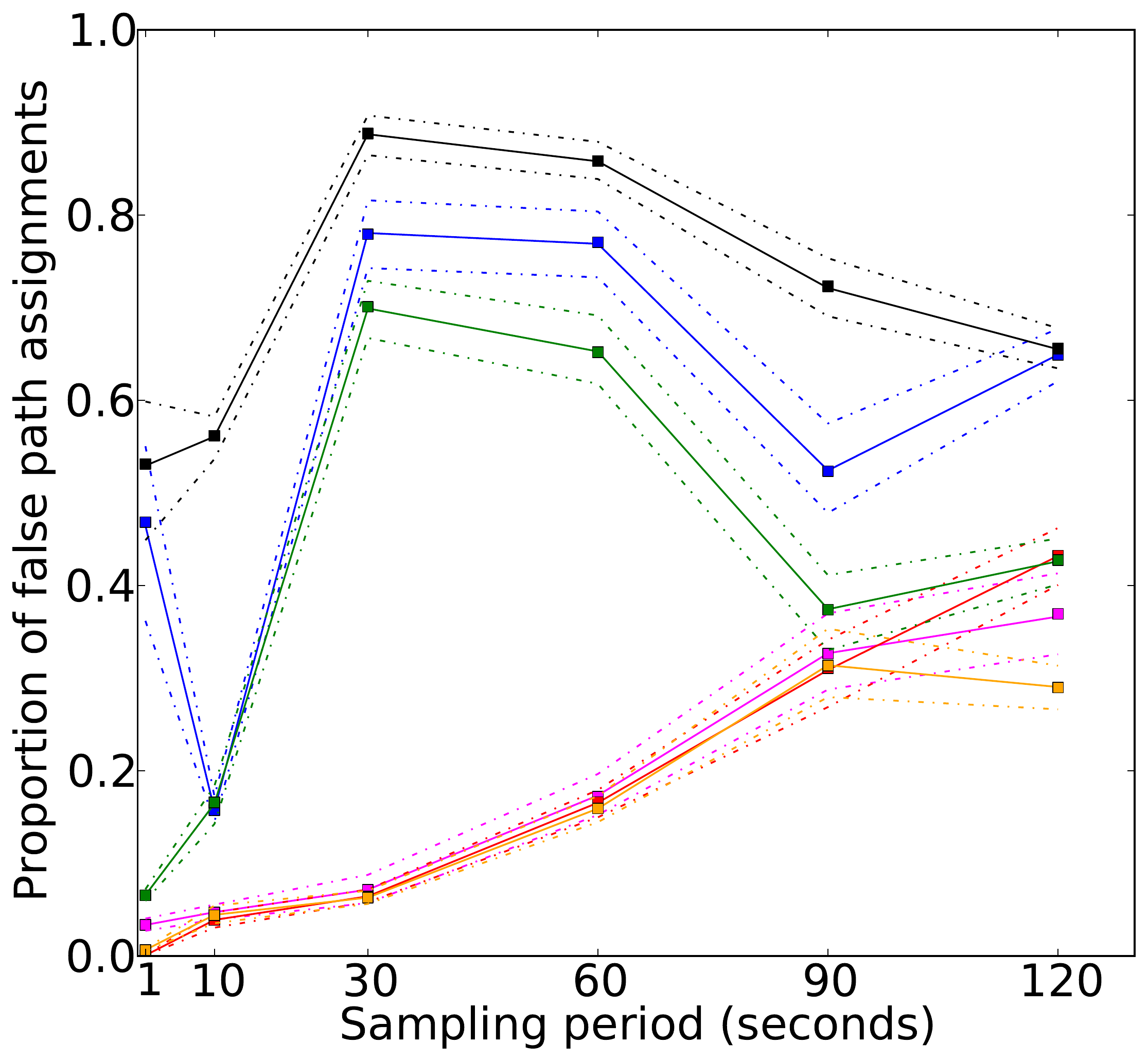}
\par\end{centering}

\caption{\label{Flo:path_miss}Path misses using the Viterbi reconstruction
for different models and different sampling rates, as a percentage
on each trajectory (lower is better). The solid line denotes the median,
the squares denote the mean and the dashed lines denote the 98\% percentiles.
The error rate is close to 0 for high frequencies: the paths are correctly
identified. In higher sampling regions, there are many more paths
to consider and the error increases substantially. Nevertheless, the
probabilistic models still perform very well: even at 2 minute intervals,
they are able to recover about 75\% of the true paths. In particular,
in these regions the shortest path becomes a viable choice for most
paths. Note how the greedy path reconstruction fails rapidly as the
sampling increases. Also note how the shortest path heuristic performs
poorly.}
\end{figure}

The most important practical result is the raw accuracy of the filter:
for each trajectory, which proportion of the paths or of the points
was correctly identified? These results are presented in Figure \ref{Flo:point_miss}
and Figure \ref{Flo:path_miss}. As expected, the error rate is 0
for high frequencies (low sampling period): all the points are correctly
identified by all the algorithms. In the low frequencies (high sampling
periods), the error is still low (around 10\%) for the trained models,
and also for the greedy model ({}``Hard closest point''). For sampling
rates between 10 seconds and 90 seconds, trained models ({}``Simple''
and {}``Complex'') show a much higher performance compared to untrained
models ({}``Hard closest point'', {}``Closest point'' and {}``Shortest
path'').

\begin{figure}
\begin{centering}
\includegraphics[width=3in]{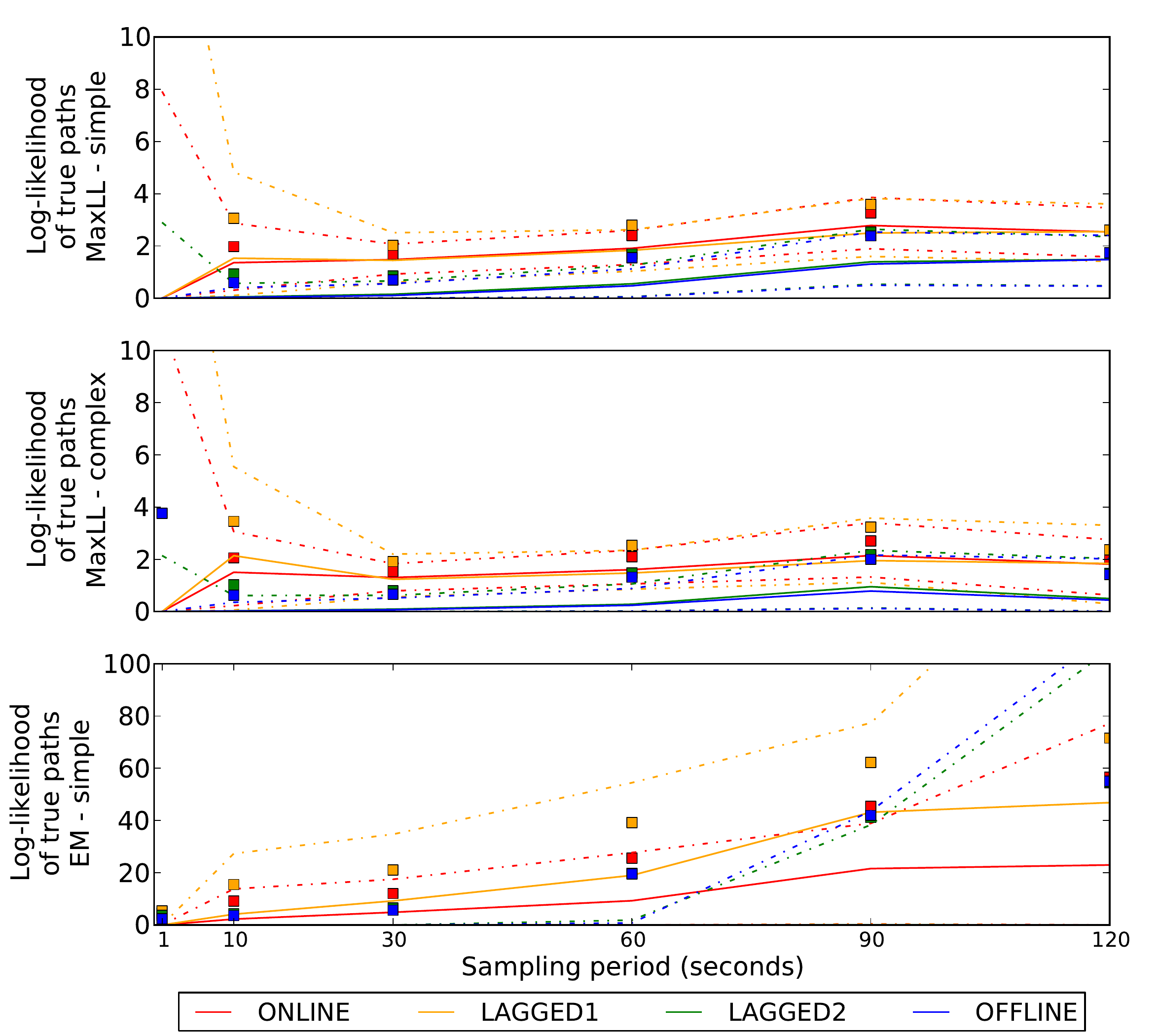}
\par\end{centering}

\caption{\label{Flo:path_ll}(Negative of) Log likelihood of true paths for
different strategies and different sampling rates (positive, lower
is better). The error bars denote the first and last quartiles (the
25th and 75th percentiles). The solid line denotes the median, the
squares denote the mean and the dashed lines denote the 98\% confidence
interval. The likelihood decreases as the sampling interval increases,
which was to be expected. Note the relatively high mean likelihood
compared to the median : a number of true paths are assigned very
low likelihood by the model, but this phenomenon is mitigated by using
better filtering strategies (2-lagged and smoothing). The use of a
more complex model (that accounts for a finer set of features for
each path) brings some improvements on the order of 25\% of all metrics.
The behavior around high frequencies (1 and 10 second time intervals)
is also very interesting. Most of the paths are chosen nearly perfectly
(the median is 0), but the filters are generally too confident and
assign very low probabilities to their outputs, which is why the likelihood
has a very heavy tail at high frequency. Note also that in the case
of high frequency, the use of an offline filter brings significantly
more accurate results than a 2-lagged filter. This difference disappears
rapidly (it becomes insignificant at 10 second intervals). Note how
the EM trained filter performs worse in the low frequencies (note
the difference of scale). The points for online strategy (red) and
for 2-lagged filtering (green) do not appear because they are too
close to the 1-lagged and offline strategies, respectively. Again
in the EM setting, the offline and 2-lagged filters perform considerably
better than the cruder strategies.}
\end{figure}

\begin{figure}
\begin{centering}
\includegraphics[width=3in]{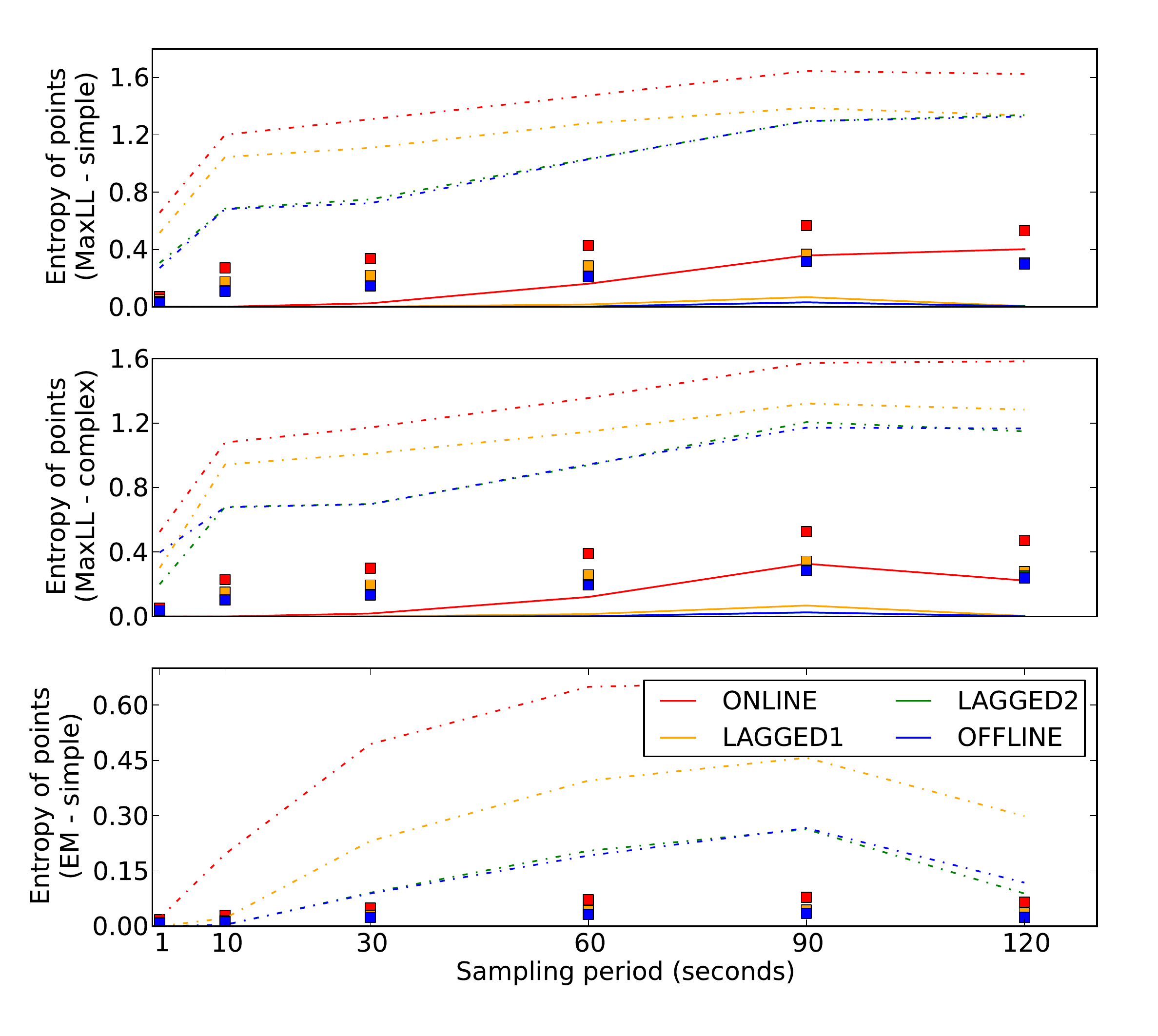}
\par\end{centering}

\caption{\label{Flo:point_entropy}Distributions of point entropies with respect
to sampling and for different models. The colors show the performance
of different filtering strategies (pure online, 1-lag, 2-lag and offline).
The entropy is a measure of the confidence of the filter on its output
and quantifies the spread of the probability distribution over all
the candidate points. The solid line denotes the median, the squares
denote the mean and the dashed lines denote the 95\% confidence interval.
The entropy starts at nearly zero for high frequency sampling : the
filters are very confident in their outputs. As sampling time increases,
the entropy at the output of the online filter increases notably.
Since the online filter cannot go back to update its belief, it is
limited to pure forward prediction and as such cannot confidently
choose a trajectory that would work in all settings. For the other
filtering strategies, the median is close to zero while the mean is
substantially higher. Indeed, the filter is very confident in its
output most of the time and assigns a weight of nearly one to one
candidate, and nearly zero to all the other outputs, but it is uncertain
in a few cases. These few cases are at the origin of the fat tail
of the distributions of entropies and the relatively wide confidence
intervals. Note that using a more complex model improves the mean
entropy by about 15\%. Also, in the case of EM, the entropy is very
low (note the difference of scale): the EM model is overconfident
in its predictions and tends to assigns very large weights to a single
choice, even if it not the good one.}
\end{figure}

\begin{figure}
\begin{centering}
\includegraphics[width=3in]{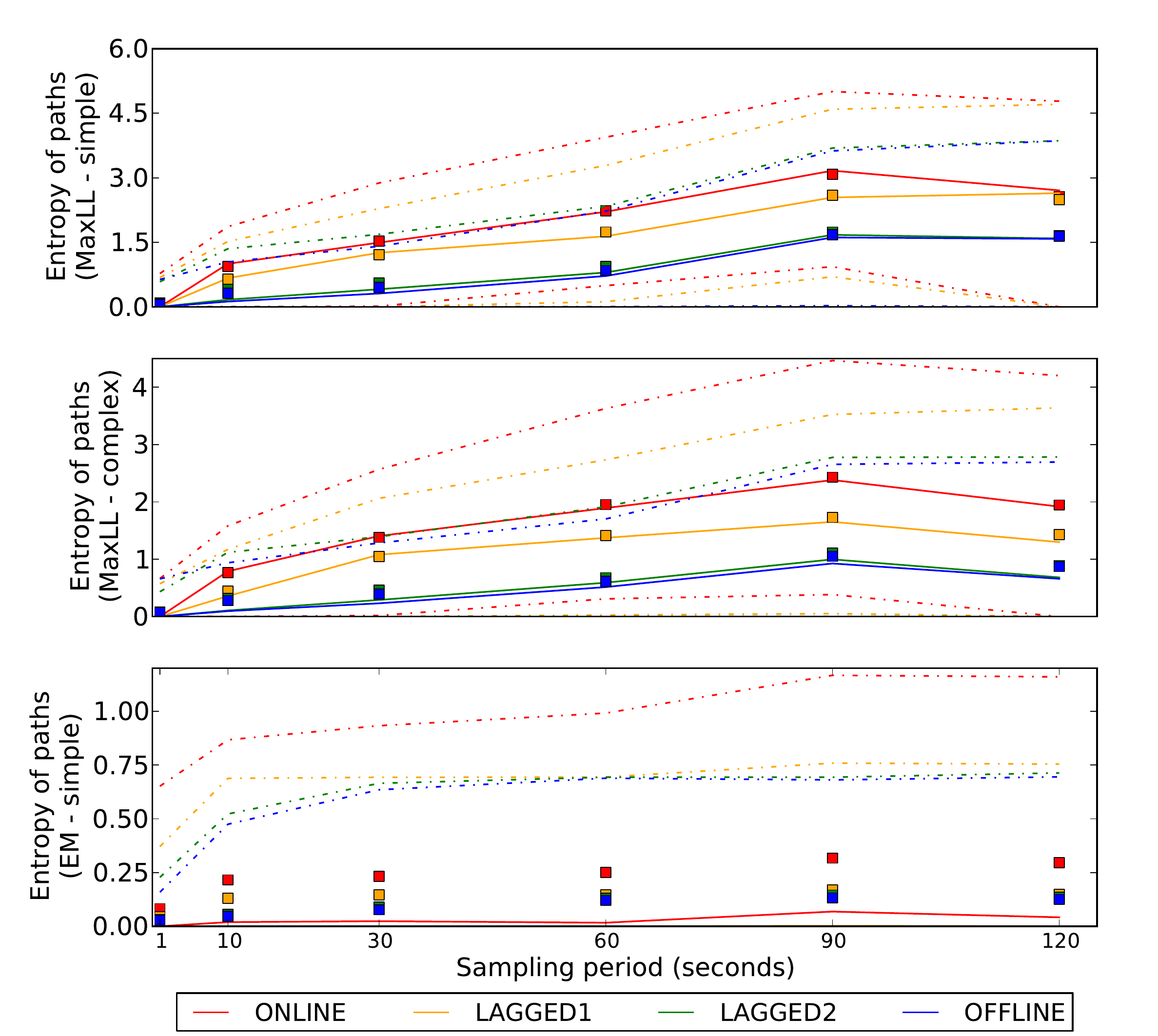}
\par\end{centering}

\caption{\label{Flo:path_entropy}Distributions of path entropies with respect
to sampling period and for different models (positive, lower is better).
The colors show the performance of different filtering strategies
(purely online, 1-lag, 2-lag and offline) The entropy is a measure
of the confidence of the filter on its output and quantifies the spread
of the probability distribution over all the candidate paths. The
solid line denotes the median, the squares denote the mean and the
dashed lines denote the 95\% confidence interval. Compared to the
points, the paths distributions have a higher entropy: the filter
is much less confident in choosing a single path and spreads the probability
weights across several choices. Again, the use of 2-lagged smoothing
is as good as pure offline smoothing, for the same computing cost
and a fraction of the data. Online and 1-lagged smoothing perform
about as well, and definitely worse than 2-lagged smoothing. The use
of a more complex model strongly improves the performance of the filter:
it results in more compact distribution over candidate paths. Again,
the model learned with EM is overconfident and tends to offer favor
a single choice, except for a few path distributions.}
\end{figure}

We now turn our attention to the resilience of the models, i.e. how
they perform when they make mistakes. We use two statistical measures:
the (log) likelihood of the true paths (Figure \ref{Flo:path_ll})
and the entropy of the distribution of points or paths (Figures \ref{Flo:point_entropy}
and \ref{Flo:path_entropy}). Note that in a perfect reconstruction
with no ambiguity, the log likelihood would be zero. Interestingly,
the log likelihoods appear very stable as the sampling interval grows:
our algorithm will continue to assign high probabilities to the true
projections even when many more paths can be used to travel from one
point to the other. The performance of the simple and the complex
models improves greatly when some backward filtering steps are used,
and stays relatively even across different time intervals. 

\begin{figure}
\begin{centering}
\includegraphics[width=3in]{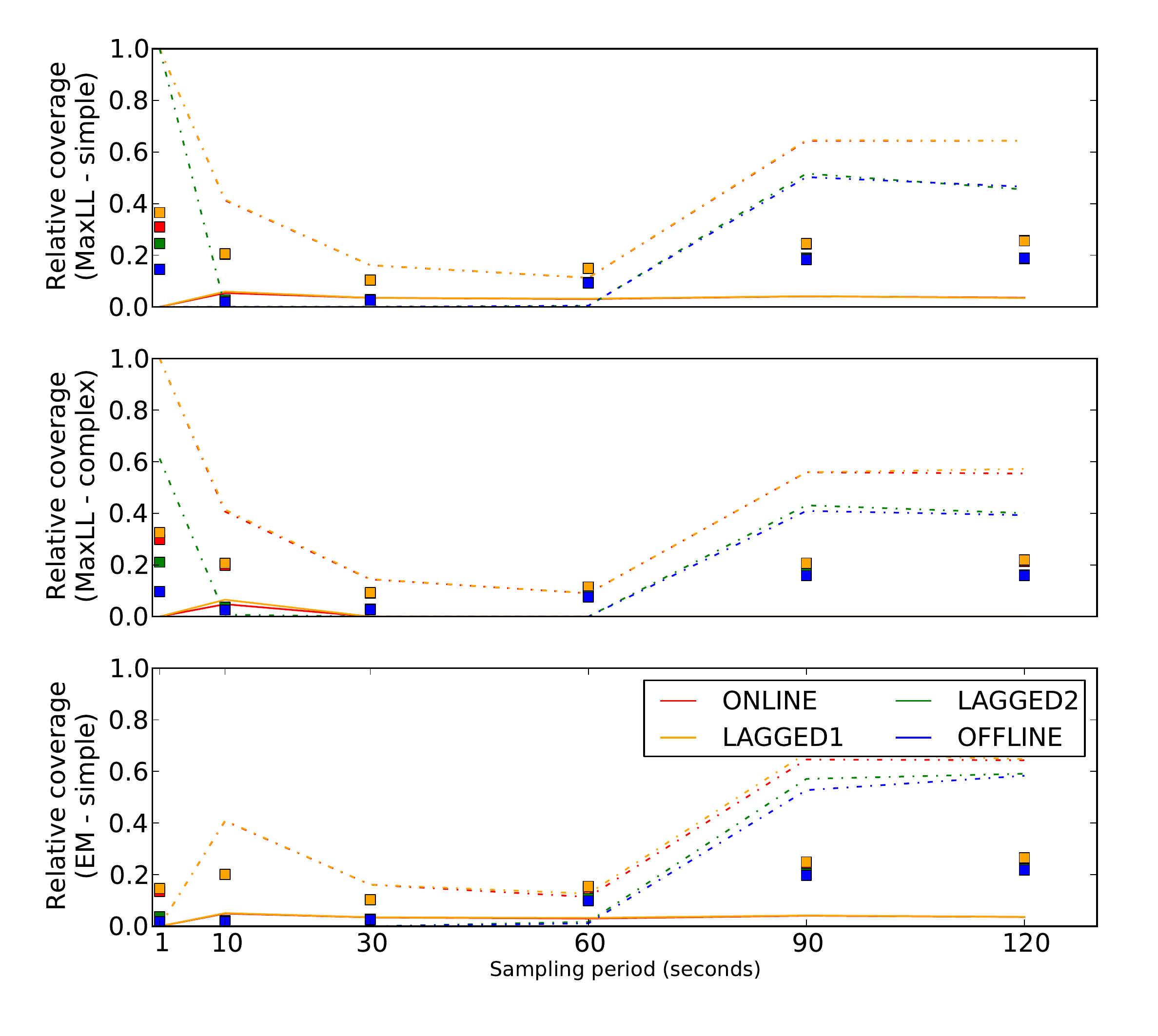}
\par\end{centering}

\caption{\label{Flo:relative-coverage-paths}Distribution of relative miscoverage
of the paths (between 0 and 1, lower is better). The solid line denotes
the median, the squares denote the mean and the dashed lines denote
the 98\% confidence interval. This metric evaluates how much of the
true path the most likely path covers , with respect to length (0
if it is completely different, 1 if the two paths overlap completely).
Two groups clearly emerge as far as computing strategies are concerned:
the online/1-lag group (orange and red curves) and the 2-lag and offline
group (green and blue curves). The relative miscoverage for the latter
group is so low that more than half of the mass is at the 0 and cannot
be seen on the curve. There are still a number of outliers that raise
the curve of the last quartile as well as the mean, especially in
the lower frequencies. Note that the paths offered by the filter are
never dramatically different: at two minute time intervals (for which
the paths are 1.7km on average), the returned path spans more than
80\% of the true path on average. The use of a more complicated model
decreases the mean miscoverage as well as the quartile metrics by
more than 15\%. Note that there is a large spread of values at high
frequency: indeed the metric is based on length covered and at high
frequency, the vehicle may be stopped and cover 0 length. This metric
is thus less indicative at high frequency.}
\end{figure}

We conclude the performance analysis by a discussion of the miscoverage
(Figure \ref{Flo:relative-coverage-paths}). The miscoverage gives
a good indication of how far the path chosen by the filter differs
from the true path. Even if the paths are not exactly the same, some
very similar path may get selected, that may differ by a turn around
a block. Note that the metric is based on length covered. At high
frequency however, the vehicle may be stopped and cover a length 0.
This metric is thus less useful at high frequency. A more complex
model improves the coverage by about 15\% in smoothing. In high sampling
resolution, the error is close to zero: the paths considered by the
filter, even if they do not match perfectly, are very close to the
true trajectory for lower frequencies. Two groups clearly emerge as
far as computing strategies are concerned: the online/1-lag group
(orange and red curves) and the 2-lag and offline group (green and
blue curves). The relative miscoverage for the latter group is so
low that more than half of the probability mass is at zero. A number
of outliers still raise the curve of the last quartile as well as
the mean, especially in the lower frequencies. The paths inferred
by the filter are never dramatically different: at two minute time
intervals (for which the paths are 1.7km on average), the returned
path spans more than 80\% of the true path on average. The use of
a more complicated model decreases the mean miscoverage as well as
all quartile metrics by more than 15\%.

\begin{figure}
\hfill{}\includegraphics[width=0.9\columnwidth]{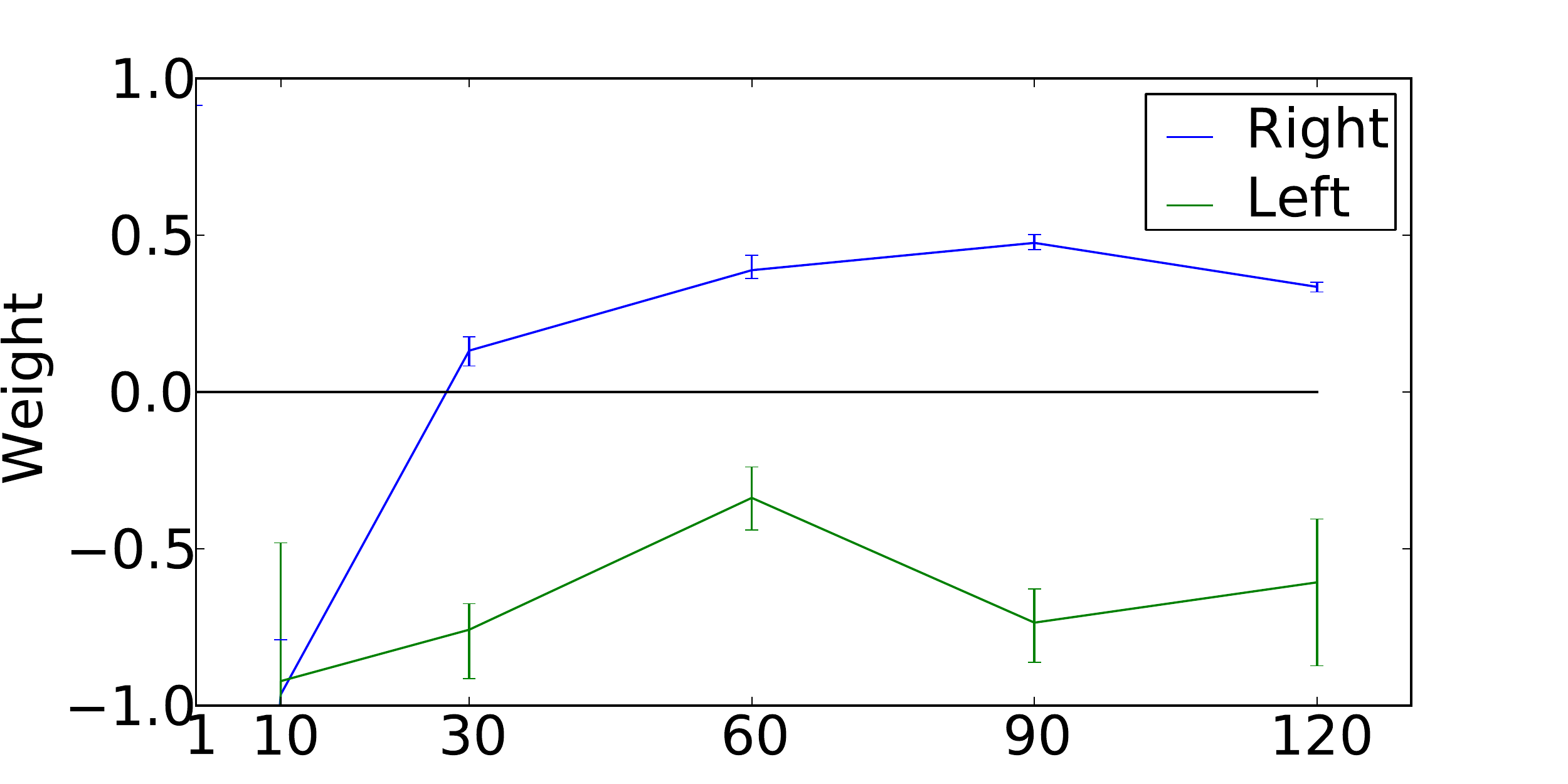}\hfill{}

\caption{\label{fig:left-right}Learned weights for left or right turns preferences.
The error bars indicate the complete span of values computed for each
time (0th and 100th percentile). For small time intervals, any turning
gets penalized but rapidly the model learns how to favor paths with
right turns against paths with left turns. A positive weight even
means that - all other factors being equal! - the driver would prefer
turning on the right than going straight.}
\end{figure}

In the case of the complex model, the weights can provide some insight
into the features involved in the decision-making process of the driver.
In particular, for extended sampling rates (t=120s), some interesting
patterns appear. For example, the drivers do not show a preference
between driving through stop signs ($w_{3}=-0.24\pm0.07$) or through
signals ($w_{4}=-0.21\pm0.11$). However, drivers show a clear preference
to turn on the right as opposed to the left, as seen in Figure \ref{fig:left-right}.
This is may be attributed, in part, to the difficulty in crossing
an intersection in the United States.

\begin{figure}
\begin{centering}
\includegraphics[width=3in]{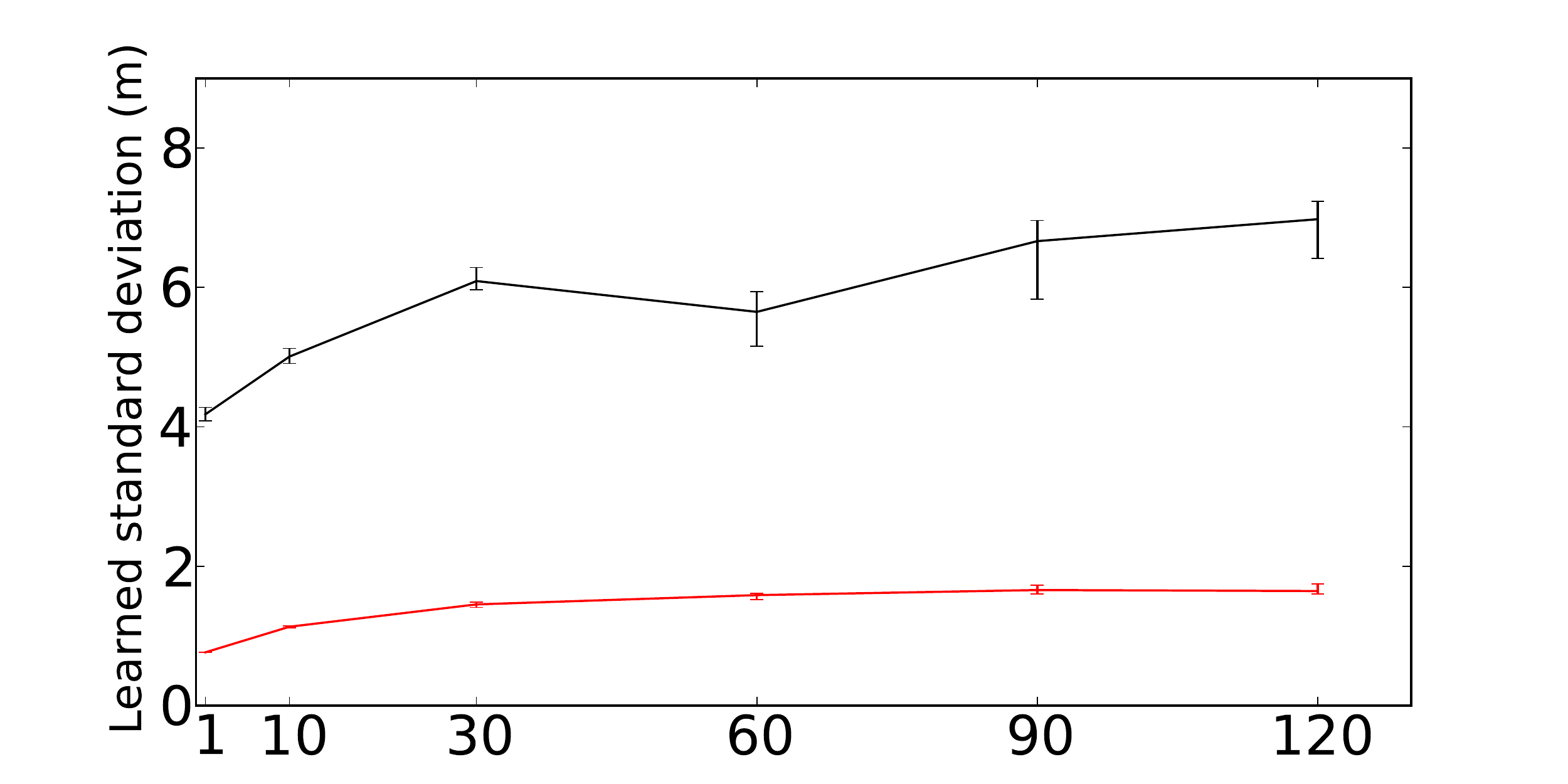}
\par\end{centering}

\caption{Standard deviation learned by the simple models, in the supervised
(Maximum Likelihood) setting and the EM setting. The error bars indicate
the complete span of values computed for each time. Note that the
maximum likelihood estimator rapidly converges toward a fixed value
of about 6 meters across any sampling time. The EM procedure also
rapidly converges, but it is overconfident and assigns a lower standard
deviation overall.}

\end{figure}

\begin{figure}
\begin{centering}
\includegraphics[width=3in]{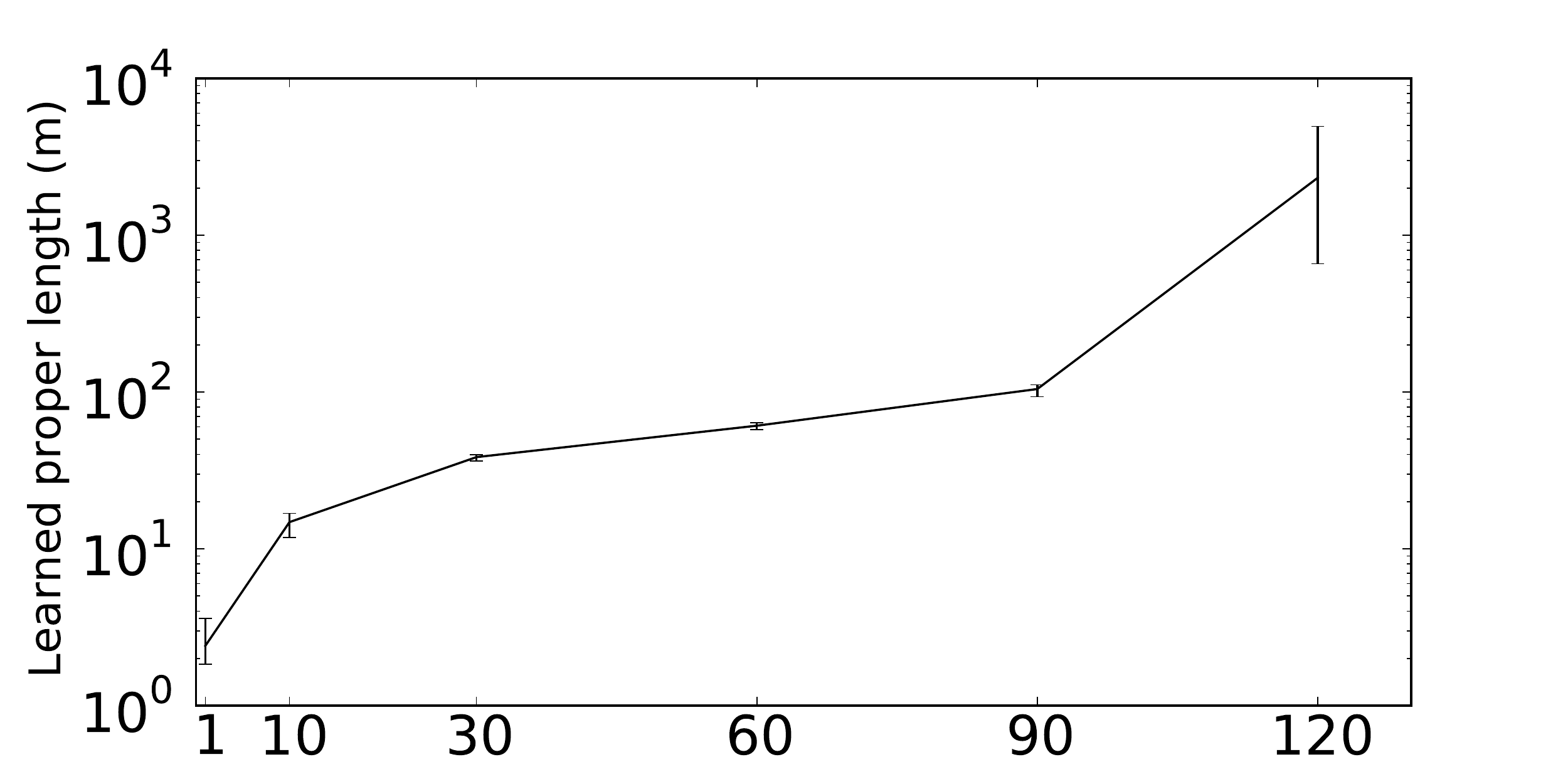}
\par\end{centering}

\caption{Characteristic length learned by the simple models, in the supervised
(Maximum Likelihood) setting and the EM setting. As hoped, it roughly
corresponds to the expected path length. The error bars indicate the
complete span of values computed for each time (0th and 100th percentile).
Note how the spread increases for large time intervals. Indeed, vehicles
have different travel lengths at such time intervals, ranging from
nearly 0 (when waiting at a signal) to more than 3 km (on the highway)
and the models struggle to accommodate a single characteristic length.
This justifies the use of more complicated models.}
\end{figure}

From a computation perspective, given a driver model, the filtering
algorithm can be dramatically improved for about as much computations
by using a full backward-forward (smoothing) filter. Smoothing requires
backing up an arbitrary sequence of points while 2-lagged smoothing
only requires the last two points. For a slightly greater computing
cost, the filter can offer a solution with a lag of one or two interval
time units that is very close to the full smoothing solution. Fixed-lag
smoothing will be the recommended solution for practical applications,
as it strikes a good balance of computation costs, accuracy and timeliness
of the results.

It should be noted the algorithm continues to provide decent results
even when points grow further apart. The errors steadily increase
with the sampling rate until the 30 seconds time interval, after which
most metrics reach some plateau. This algorithm could be used in tracking
solutions to improve the battery life of the device by up to an order
of magnitude for GPSs that do not need extensive warm up. In particular,
the tracking devices of fleet vehicle are usually designed to emit
every minute as the road-level accuracy is not a concern in most cases.

\subsection{Unsupervised learning results}

The filter was also trained for the simple and complex models using
Dataset 2. This dataset does not include true observations but is
two orders of magnitude larger than Dataset 1 for the matching sampling
period (1 minute). We report some comparisons between the models previously
trained with Dataset 1 ({}``MaxLL-Simple'', {}``EM-Simple'', {}``MaxLL-Complex'')
and the same simple and complex models trained on Dataset 2: {}``EM-Simple
large'' and {}``EM-Complex large''. The learning procedure was
calibrated using cross-validation and was run in the following way:
all unsupervised models were initialized with a hand-tuned heuristic
model involving only the standard deviation and the characteristic
length (with the weight of all the features set to 0). The Expectation-Maximization
algorithm was then run for 3 iterations. Inside each EM iteration,
the M-step was run with a single Newton-Raphson iteration at each
time, using the full gradient and Hessian and a quadratic penalty
of $10^{-2}$. During the E step, each sweep over the data took 13
hours 400 thousand points on a 32-core Intel Xeon server.

We limit our discussion to the main findings for brevity. The unsupervised
training finds some weight values similar to those found with supervised
learning. The magnitude of these weights is larger than in the supervised
settings. Indeed, during the E step, the algorithm is free to assign
any sensible value to the choice of the path. This may lead to a self-reinforcing
behavior and the exploration of a bad local minimum. 

\begin{figure}
\begin{centering}
\includegraphics[width=0.9\columnwidth]{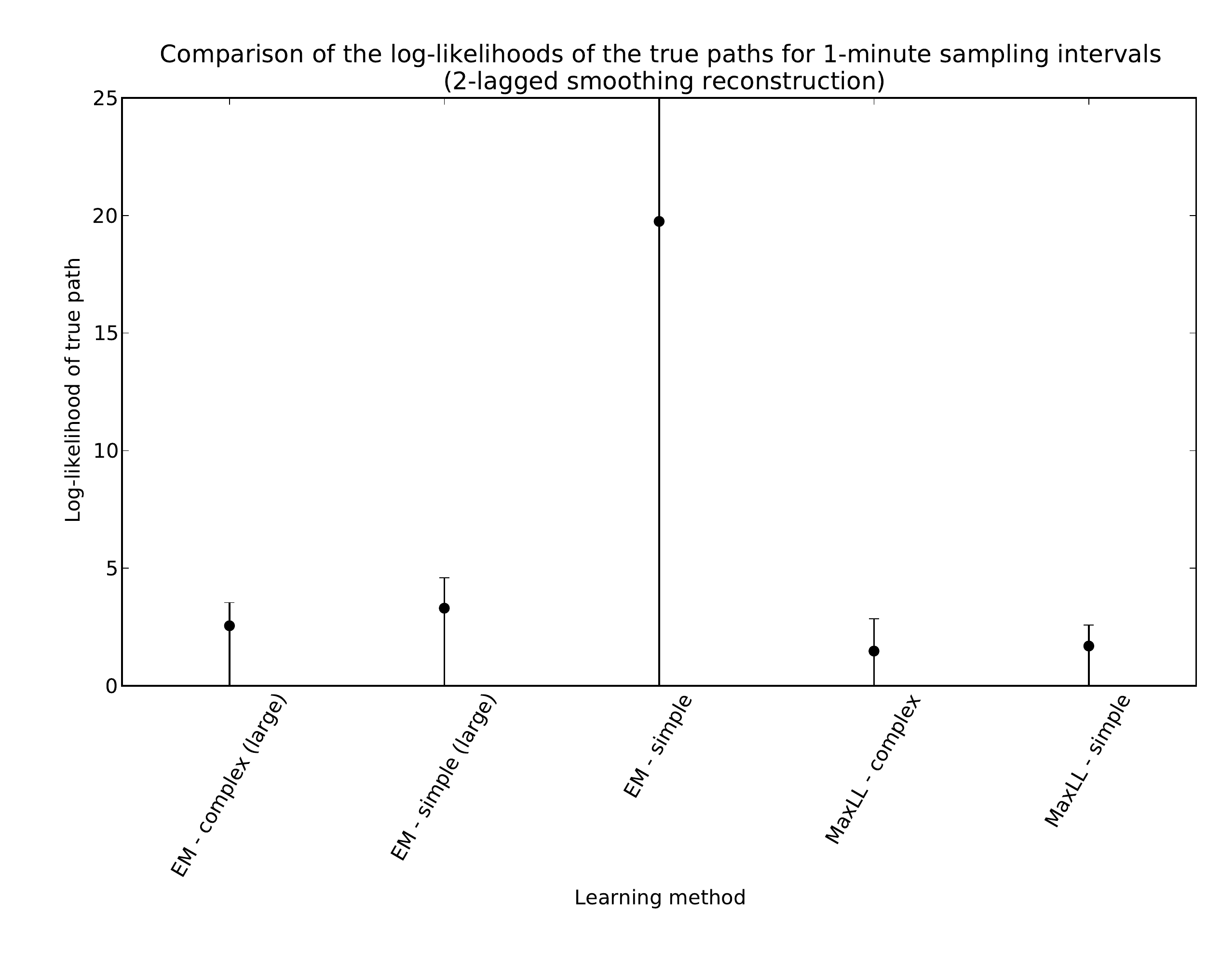}
\par\end{centering}

\caption{Expected likelihood of the true path. The central point is the mean
log-likelihood, the error bars indicate the 70\% confidence interval.
Note that the simple model trained unsupervised with the small dataset
has a much larger error, i.e. it assigns low probabilities to the
true path. Both unsupervised models tend to express the same behavior
but are much more robust.\label{fig:em_ll_paths}}
\end{figure}

\begin{figure}
\begin{centering}
\includegraphics[width=0.9\columnwidth]{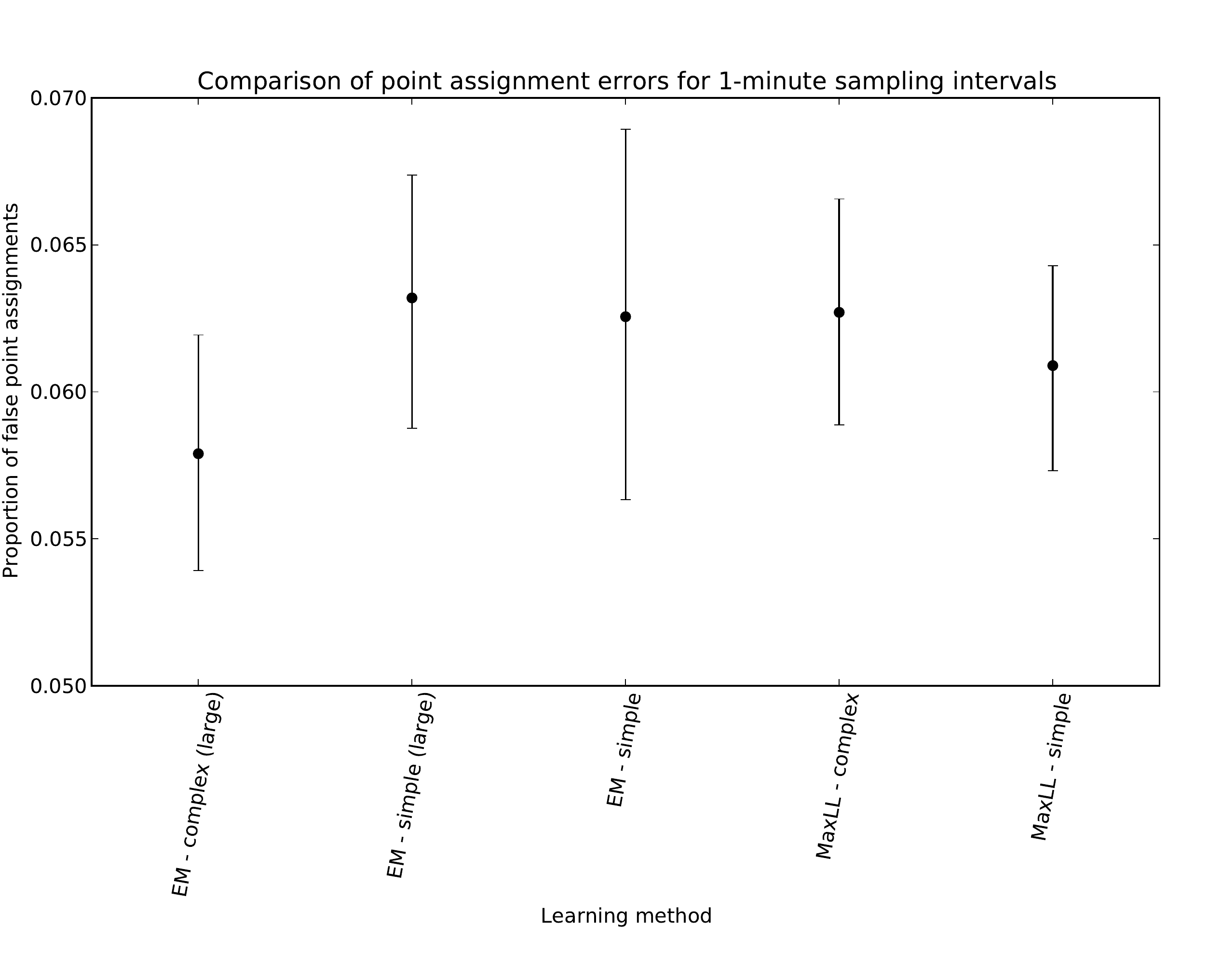}
\par\end{centering}

\caption{Proportion of true points incorrectly identified, for different models
evaluated with 1-minute sampling (lower is better). The central point
is the mean proportion, the error bars indicate the 70\% confidence
interval. Unsupervised models are very competitive against supervised
models, and the complex unsupervised model slightly outperforms all
supervised models.\label{fig:em_true_points_percentage}}
\end{figure}

\begin{figure}
\begin{centering}
\includegraphics[width=0.9\columnwidth]{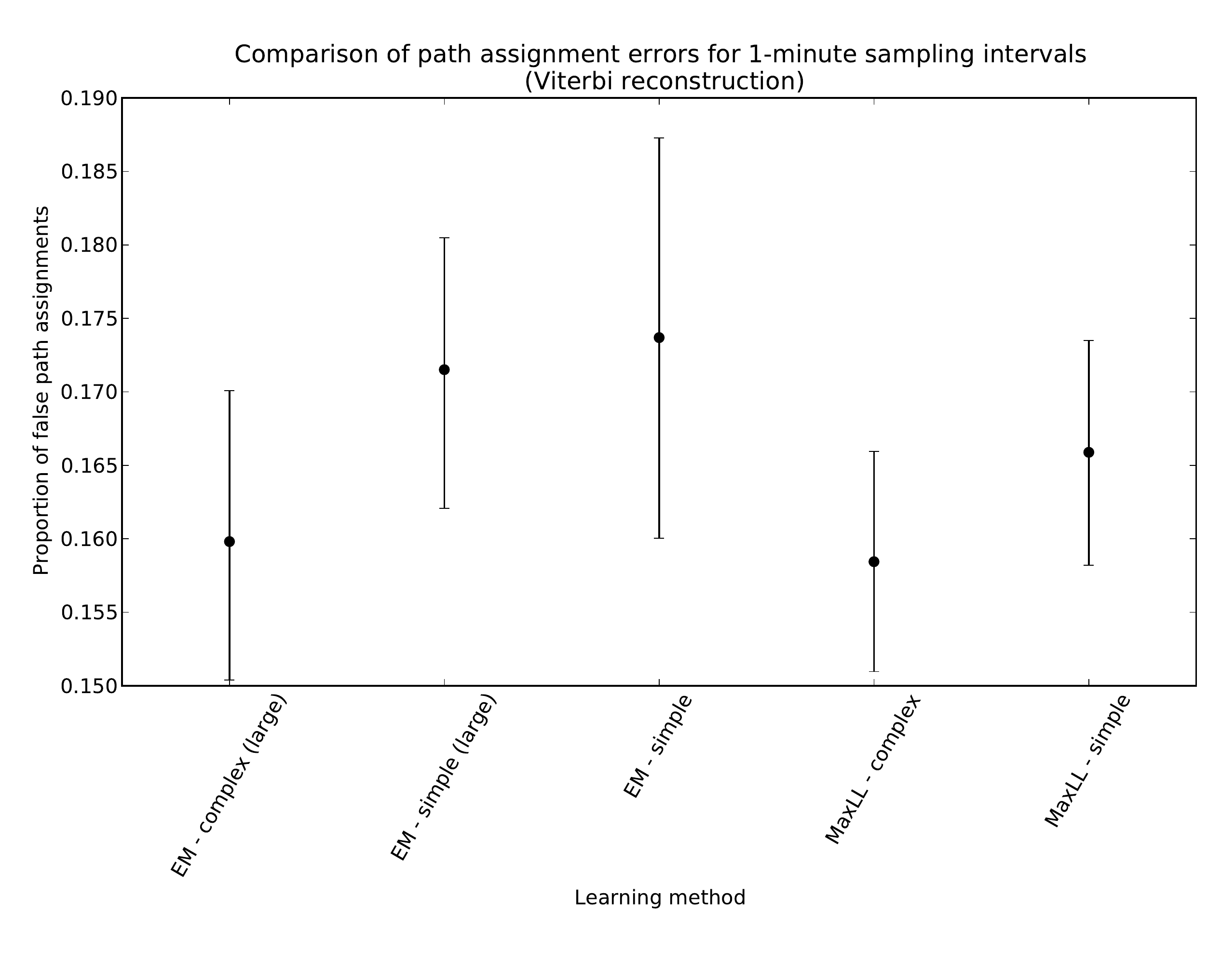}
\par\end{centering}

\caption{Proportion of true paths incorrectly identified, for different models
evaluated with 1-minute sampling (lower is better). The central point
is the mean proportion, the error bars indicate the 70\% confidence
interval. The complex unsupervised model is as good as the best supervised
model.\label{fig:em_true_paths_percentage}}
\end{figure}

As Figures \ref{fig:em_ll_paths}, \ref{fig:em_true_paths_percentage},
and \ref{fig:em_true_points_percentage} show, a large training dataset
puts unsupervised methods on par with supervised methods as far as
performance metrics are concerned. Also, the inspection of the parameters
learned on this dataset corroborates the finding made earlier. One
is tempted to conclude that given enough observations, there no need
to collect expensive high-frequency data to train a model.

\subsection{Key findings}

Our algorithm can reconstruct a sensible approximation of the trajectory
followed by the vehicles analyzed, even in complex urban environments.
In particular, the following conclusions can be made:
\begin{itemize}
\item An intuitive deterministic heuristic ({}``Hard closest point'')
dramatically fails for paths at low frequencies, less so for points.
It should not be considered for sampling intervals larger than 30
seconds.
\item A simple probabilistic heuristic ({}``closest point'') gives good
results for either very low frequencies (2 minutes) or very high frequencies
(a few seconds) with more 75\% of paths and 94\% points correctly
identified. However, the incorrect values are not as close to the
true trajectory as they are with more accurate models ({}``Simple''
and {}``Complex'').
\item For the medium range (10 seconds to 90 seconds), trained models (either
supervised or unsupervised) have a greatly improved accuracy compared
to untrained models, with 80\% to 95\% of the paths correctly identified
by the former.
\item For the paths that are incorrectly identified, trained models ({}``Simple''
or {}``Complex'') provide better results compared to untrained models
(the output paths are closer to the true paths, and the uncertainty
about which paths may have been taken is much reduced). Furthermore,
using a complex model ({}``Complex'') improves these results even
more by a factor of 13-20\% on all metrics.
\item For filtering strategies: online filtering gives the worst results
and its performance is very similar to 1-lagged smoothing. The slower
strategies (2-lagged smoothing and offline) outperform the other two
by far. Two-lagged smoothing is nearly as good as offline smoothing,
except in very high frequencies (less than 2 second sampling) for
which smoothing clearly provides better results.
\item Using a trained algorithm in a purely unsupervised fashion provides
an accuracy as good as when training in a supervised setting - within
some limits and assuming enough data is available. The model produced
by EM ({}``EM-Simple'') is equally good in terms of raw performance
(path and point misses) but it may be overconfident.
\item With more complex models, the filter can be used to infer some interesting
patterns about the behavior of the drivers.
\end{itemize}

\section{Conclusions and future work}

\label{sec:results}

We have presented a novel class of algorithms to track moving vehicles
on a road network: the \emph{path inference filter}. This algorithm
first projects the raw points onto candidate projections on the road
network and then builds candidate trajectories to link these candidate
points. An observation model and a driver model are then combined
in a Conditional Random Field to find the most probable trajectories.

The algorithm exhibits robustness to noise as well as to the peculiarities
of driving in urban road networks. It is competitive over a wide range
of sampling rates (1 seconds to 2 minutes) and greatly outperforms
intuitive deterministic algorithms. Furthermore, given a set of ground
truth data, the filter can be automatically tuned using a fast supervised
learning procedure. Alternatively, using enough regular GPS data with
no ground truth, it can be trained using unsupervised learning. Experimental
results show that the unsupervised learning procedure compares favorably
against learning from ground truth data. One may conclude that given
enough observations, there no need to collect expensive high-frequency
data to train a model.

This algorithm supports a range of trade-offs between accuracy, timeliness
and computing needs. In its most accurate settings, it extends the
current state of the art \cite{zheng2011weight,yuan2010interactive}.
This result is supported by the theoretical foundations of Conditional
Random Fields. Because no standardized benchmark exists, the authors
have released an open-source implementation of the filter to foster
comparison with other methodologies using other datasets \cite{pythonimpl}.

In conjunction with careful engineering, this program can achieve
high map-matching throughput. The authors have written an industrial-strength
version in the Scala programming language, deployed in the \emph{Mobile
Millennium }system. This version maps GPS points at a rate of about
400 points per second on a single core for the San Francisco Bay area
(several hundreds of thousands of road links), and has been scaled
to multicore architecture to achieve an average throughput of several
thousand points per second \cite{hunter2011SOCC}.

A number of extensions could be considered to the core framework.
In particular, more detailed models of the driver behavior as well
as algorithms for automatic feature selection should bring additional
improvements in performance. Another line of research is the mapping
of very sparse data (sampling intervals longer than two minutes).
Although the filter already attempts to consider as few trajectories
as possible, more aggressive pruning may be necessary in order to
achieve good performance. Finally, the EM procedure presented for
automatically tuning the algorithm requires large amounts of data
to be effective, and could be tested on larger datasets that what
we have presented here.

\subsection*{Authors}

\noindent \begin{wrapfigure}{i}{2cm}%
\vspace*{-.5cm}\centering\hspace*{-.5cm}\includegraphics[width=2cm]{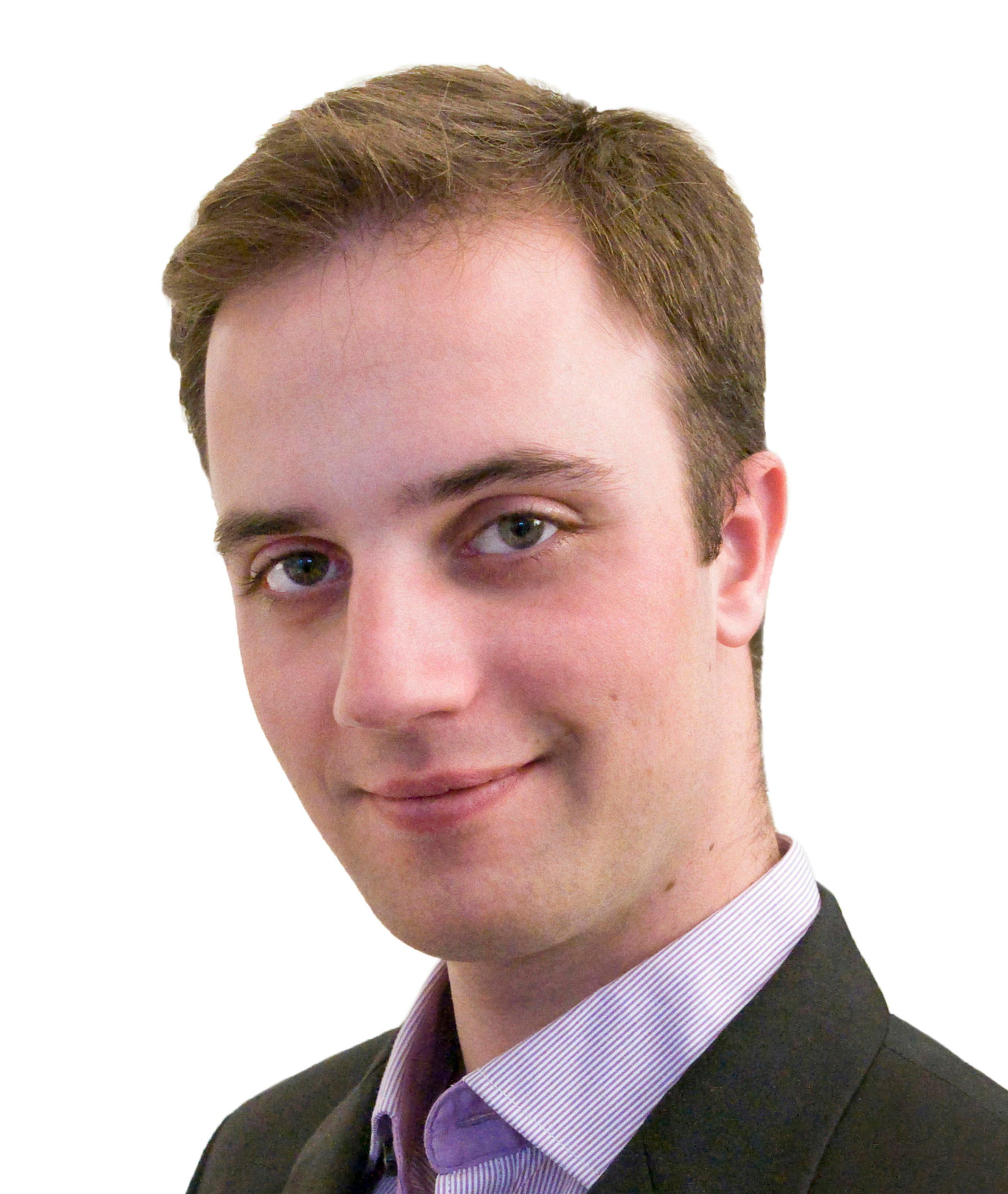}\hspace*{-.5cm}\vspace*{-.5cm}\end{wrapfigure}%
\textbf{Timothy Hunter} is a Ph.D. student in the Department of Electrical
Engineering and Computer Science and in the AMPLab, at the University
of California at Berkeley. He received the Engineering Degree in Applied
Mathematics from the Ecole Polytechnique, France, in July 2007, and
the M.S. degree in Electrical Engineering from Stanford University
in 2009. His research interests include new programming models for
Machine Learning with Big Data, and some applications to estimation
and transportation.

\noindent \begin{wrapfigure}{i}{2cm}%
\vspace*{-.5cm}\centering\hspace*{-.5cm}\includegraphics[width=2cm]{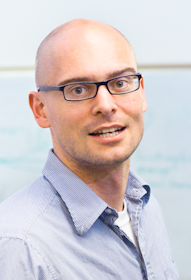}\hspace*{-.5cm}\vspace*{-.5cm}\end{wrapfigure}%
\textbf{Pieter Abbeel} received a BS/MS in Electrical Engineering
from KU Leuven (Belgium) and received his Ph.D. degree in Computer
Science from Stanford University in 2008. He joined the faculty at
UC Berkeley in Fall 2008, with an appointment in the Department of
Electrical Engineering and Computer Sciences. He has won various awards,
including best paper awards at ICML and ICRA, the Sloan Fellowship,
the Okawa Foundation award, and 2011's TR35. He has developed apprenticeship
learning algorithms which have enabled advanced helicopter aerobatics,
including maneuvers such as tic-tocs, chaos and auto-rotation, which
only exceptional human pilots can perform. His group has also enabled
the first end-to-end completion of reliably picking up a crumpled
laundry article and folding it. His work has been featured in many
popular press outlets, including BBC, MIT Technology Review, Discovery
Channel, SmartPlanet and Wired.

\noindent \begin{wrapfigure}{i}{2cm}%
\vspace*{-.3cm}\centering\hspace*{-.5cm}\includegraphics[width=2cm]{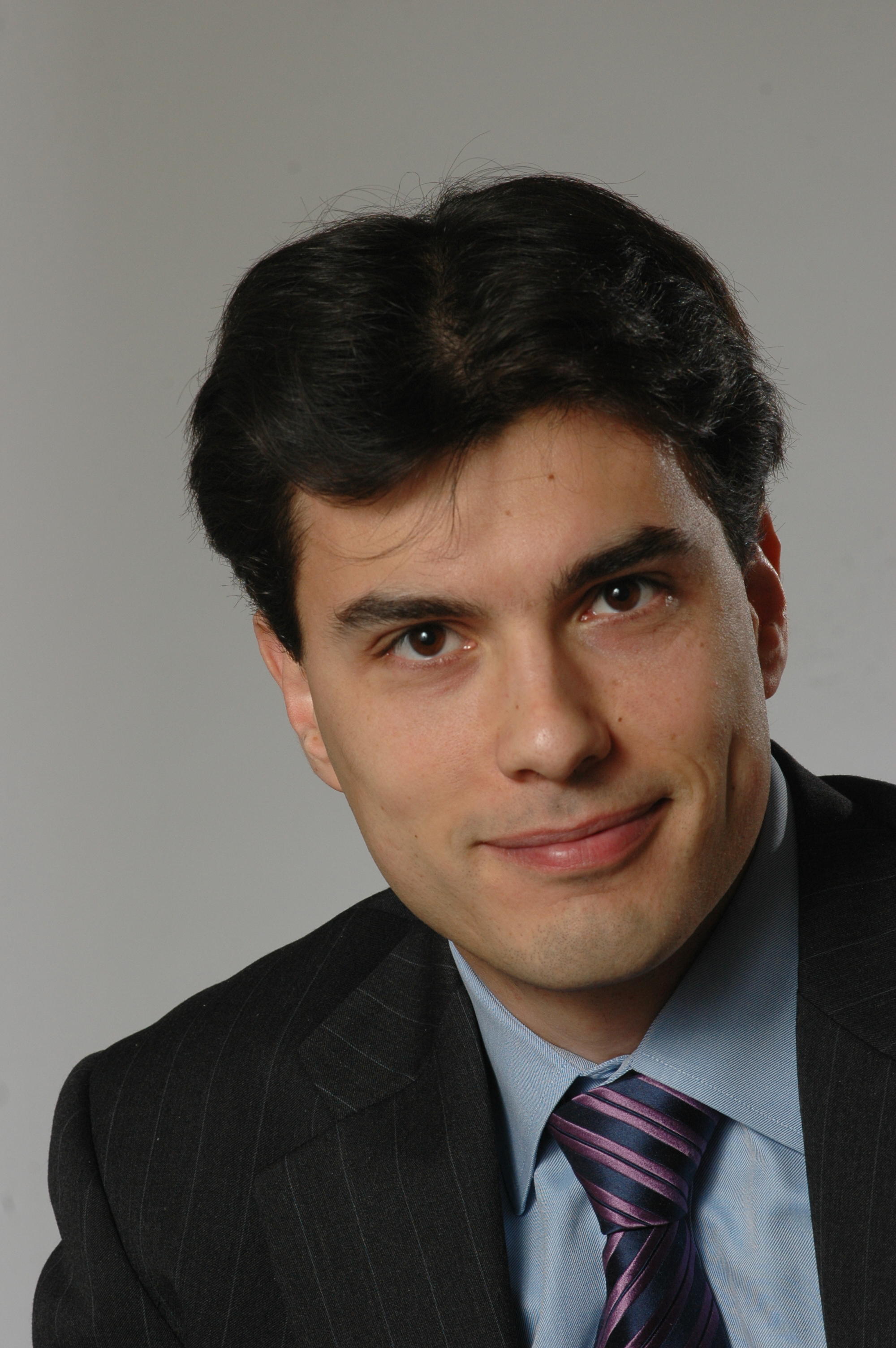}\hspace*{-.5cm}\vspace*{-.3cm}\end{wrapfigure}%
\textbf{Alexandre Bayen} received the Engineering Degree in applied
mathematics from the Ecole Polytechnique, France, in July 1998, the
M.S. degree in aeronautics and astronautics from Stanford University
in June 1999, and the Ph.D. in aeronautics and astronautics from Stanford
University in December 2003. He was a Visiting Researcher at NASA
Ames Research Center from 2000 to 2003. He worked as the Research
Director of the Autonomous Navigation Laboratory at the Laboratoire
de Recherches Balistiques et Aerodynamiques, (Ministere de la Defense,
Vernon, France), where he holds the rank of Major. He is an Associate
Professor in the Electrical Engineering and Computer Sciences at UC
Berkeley. Bayen has authored one book and over 100 articles in peer
reviewed journals and conferences. He is the recipient of the Ballhaus
Award from Stanford University, 2004, of the CAREER award from the
National Science Foundation, 2009 and he is a NASA Top 10 Innovators
on Water Sustainability, 2010. His projects Mobile Century and Mobile
Millennium received the 2008 Best of ITS Award for Best Innovative
Practice, at the ITS World Congress and a TRANNY Award from the California
Transportation Foundation, 2009. Bayen is the recipient of the Presidential
Early Career Award for Scientists and Engineers (PECASE) award from
the White House, 2010. Mobile Millennium has been featured more than
100 times in the media, including TV channels and radio stations (CBS,
NBC, ABC, CNET, NPR, KGO, the BBC), and in the popular press (Wall
Street Journal, Washington Post, LA Times).

\subsection*{Acknowledgments}

The authors wish to thank the staff at the California Center for Innovative
Transportation, in particular Ryan Herring and Saneesh Apte, for their
dedicated support. Discussions with Samitha Samarayanake have been
instrumental in the writing this article. The authors are indebted
to Warren Hoburg, and Kailin Kroetz for their thorough and insightful
comments on the draft.

This research is supported in part by gifts from Google, SAP, Amazon
Web Services, Cloudera, Ericsson, Huawei, IBM, Intel, Mark Logic,
Microsoft, NEC Labs, Network Appliance, Oracle, Splunk and VMWare.
The generous support of the US Department of Transportation and the
California Department of Transportation is gratefully acknowledged.
We also thank Nokia and NAVTEQ for the ongoing partnership and support
through the \emph{Mobile Millennium} project. 
\cleardoublepage{}

\section*{Notation}

\begin{tabular}{|>{\centering}p{0.2\textwidth}|>{\raggedright}p{0.8\textwidth}|}
\hline 
Symbol & Meaning\tabularnewline
\hline
\hline 
$\underline{\delta}\left(x,p\right)$ & Compatibility function between a state $x$ and the start state of
a path $p$\tabularnewline
\hline 
$\bar{\delta}\left(p,x\right)$ & Compatibility function between an end state $x$ and the end state
of a path $p$\tabularnewline
\hline 
$\epsilon=\sigma^{-2}$ & Stacked inverse variance\tabularnewline
\hline 
$\eta=\eta\left(p|x\right)$ & Transition model\tabularnewline
\hline 
$\theta$ & Stacked vector of parameters\tabularnewline
\hline 
$\mu$ & Weight vector\tabularnewline
\hline 
$\xi_{1},\xi_{2}$ & Simple features (path length and distance of a point projection to
its GPS coordinate)\tabularnewline
\hline 
$\pi$ & Probability distribution, the variables are always indicated to disambiguate
which variables are involved\tabularnewline
\hline 
$\hat{\pi}$ & Probability distribution in the case of a dynamic Bayesian network,
the variables are always indicated to disambiguate which variables
are involved\tabularnewline
\hline 
$\widetilde{\pi}$ & Expected plug-in distribution\tabularnewline
\hline 
$\varsigma$ & Set of valid trajectories\tabularnewline
\hline 
$\sigma$ & Standard deviation\tabularnewline
\hline 
$\tau=x^{1}p^{1}x^{2}...p^{T-1}x^{T}$ & Trajectory of a vehicle\tabularnewline
\hline 
$\tau^{*}$ & Most likely trajectory given a model $\left(\omega,\eta\right)$ and
a GPS track $g^{1:T}$\tabularnewline
\hline 
$\phi\left(\tau|g^{1:T}\right)$ & Potential, or unnormalized score, of a trajectory\tabularnewline
\hline 
$\phi_{i}^{t}$ & Maximum of all the potentials of the partial trajectories that end
in the state $x_{i}^{t}$\tabularnewline
\hline 
$\phi^{*}$ & Maximum value over all the potentials of the trajectories compatible
with $g^{1:T}$\tabularnewline
\hline 
$\varphi\left(p\right)$ & Feature function\tabularnewline
\hline 
$\psi\left(z^{1:L}\right)$ & Generalized potential function\tabularnewline
\hline 
$\omega=\omega\left(g|x\right)$ & Observation model\tabularnewline
\hline 
$\Omega\left(x\right)$ & Prior distribution over the states $x$\tabularnewline
\hline 
$g$ & GPS coordinate (pair of latitude and longitude)\tabularnewline
\hline 
$\left(g^{t}\right)^{1:T}$ & Sequence of all $T$ GPS observations of a GPS track\tabularnewline
\hline 
$I^{t}$ & Number of projected states of the GPS point at time index $t$ onto
the road network\tabularnewline
\hline 
$I$ & Number of mappings of the GPS point onto the road network\tabularnewline
\hline 
$J$ & Number of all candidate trajectories between the mappings $\mathbf{x}$
and $\mathbf{x'}$\tabularnewline
\hline 
$J^{t}$ & Number of all trajectories between the mappings at time $t$ (i.e.
$\mathbf{x}^{t}$) and the mappings at time $t+1$ $\mathbf{x}^{t+1}$\tabularnewline
\hline 
$\left(l,o\right)$ & Location in the road network defined by a pair of a road link $l$
and an offset position $o$ on this link\tabularnewline
\hline 
$L=2T-1$ & Complete length of a trajectory\tabularnewline
\hline 
$\mathcal{L}$ & Expected likelihood\tabularnewline
\hline 
$\mathcal{N}=\left(\mathcal{V},\mathcal{E}\right)$ & Road network, comprising some vertices (nodes) $\mathcal{N}$ and
edges (roads) $\mathcal{E}$\tabularnewline
\hline 
$x=\left(l,o\right)$ & State of the vehicle (typically a location on the road network)\tabularnewline
\hline 
$p$ & Path between one mapping $x$ and one subsequent mapping $x'$\tabularnewline
\hline 
$\mathbf{p}=\left(p_{j}\right)_{1:J}$ & Collection of all $J$ candidate trajectories between a set of candidate
states$\mathbf{x}$ and a subsequent set $\mathbf{x}'$\tabularnewline
\hline 
$\mathbf{p}^{t}=\left(p_{j}^{t}\right)_{1:J^{t}}$ & Collection of all $J$ candidate trajectories between the set of candidate
states at time $t$ $\mathbf{x}^{t}$ and the subsequent set $\mathbf{x}^{t+1}$\tabularnewline
\hline 
$\bar{q}_{i}^{t}$ & Probability that the vehicle is in the discrete state $x_{i}^{t}$
at time $t$ given all observations\tabularnewline
\hline 
$\left.\overrightarrow{q}\right._{i}^{t}$ & Probability that the vehicle is in the discrete state $x_{i}^{t}$
at time $t$ given all observations up to time $t$\tabularnewline
\hline 
$\left.\overleftarrow{q}\right._{i}^{t}$ & Probability that the vehicle is in the discrete state $x_{i}^{t}$
at time $t$ given all observations after time $t+1$\tabularnewline
\hline 
$\overline{r}_{j}^{t}$ & Probability that the vehicle uses the (discrete) path $p_{j}^{t}$
at time $t$ given all observations\tabularnewline
\hline 
$\left.\overrightarrow{r}\right._{j}^{t}$ & Probability that the vehicle uses the (discrete) path $p_{j}^{t}$
at time $t$ given all observations up to time $t$\tabularnewline
\hline 
$\left.\overleftarrow{r}\right._{j}^{t}$ & Probability that the vehicle uses the (discrete) path $p_{j}^{t}$
at time $t$ given all observations after time $t+1$\tabularnewline
\hline 
$T$ & Number of GPS observations for a track\tabularnewline
\hline 
$T^{l}\left(z^{l}\right)$ & Generalized feature vector\tabularnewline
\hline 
$Z$ & Partition function\tabularnewline
\hline
\end{tabular}
\cleardoublepage{}\bibliographystyle{plain}
\bibliography{path_inference}

\end{document}